%% file: main.tex
\documentclass[conference]{IEEEtran}
\pdfoutput=1

\usepackage[utf8]{inputenc}

\usepackage{amsfonts}
\usepackage{amsmath}
\usepackage{booktabs}
\usepackage{multirow}
\usepackage{graphicx}
\newtheorem{theorem}{Theorem}

\usepackage{subcaption}
\usepackage{hyperref}

\input{macros}

\title{Model extraction from counterfactual explanations}
\author{Ulrich A{\"i}vodji\IEEEauthorrefmark{3},
Alexandre Bolot \IEEEauthorrefmark{4}, 
S{\'e}bastien Gambs\IEEEauthorrefmark{3}\vspace*{0.15cm} \\ 
Universit{\'e} du Qu{\'e}bec {\`a} Montr{\'e}al\IEEEauthorrefmark{3}, Polytech Nice Sophia\IEEEauthorrefmark{4} 
}

\begin{document}

\maketitle

\input{abstract}

\input{introduction}

\input{preliminaries}

\input{contributions}

\input{experiments}

\input{related}

\input{discussions}

\input{conclusion}

\section*{Acknowledgments}
\noindent  S{\'e}bastien Gambs is supported by the Canada Research Chair program, a Discovery Grant from NSERC, the Legalia project from the AUDACE program funded by the FQRNT and the project \emph{Privacy and Ethics: Understanding the Convergences and Tensions for the Responsible Development of Machine Learning} funded by the Office of the Privacy Commissioner of Canada (OPC). The opinions expressed in this paper are only the one of the authors and do not necessarily reflect those of the OPC.

\bibliographystyle{IEEEtran}
\bibliography{papers}

\end{document}

%% file: macros.tex
\newtheorem{definition}[theorem]{Definition}

\def\bbox{\mathcal{B}}

\def\param{\theta}
\def\paramspace{\Theta}

\def\model{f_\param{}}

\def\inputspace{\mathcal{X}}

\def\outputspace{\mathcal{Y}}

\def\singleinput{x}
\def\singleoutput{y}
\def \lossfunc{L}
\def\reg{\Omega}

\def\traindata{\mathcal{D}_{train}}

\def\adv{\mathcal{A}}

\def\surr{\mathcal{S}}

%% file: abstract.tex
\begin{abstract}
Post-hoc explanation techniques refer to \emph{a posteriori} methods that can be used to explain how black-box machine learning models produce their outcomes.
Among post-hoc explanation techniques, counterfactual explanations are becoming one of the most popular methods to achieve this objective.
In particular, in addition to highlighting the most important features used by the black-box model, they provide users with actionable explanations in the form of data instances that would have received a different outcome.
Nonetheless, by doing so, they also leak non-trivial information about the model itself, which raises privacy issues. 
In this work, we demonstrate how an adversary can leverage the information provided by counterfactual explanations to build high-fidelity and high-accuracy model extraction attacks. 
More precisely, our attack enables the adversary to build a faithful copy of a target model by accessing its counterfactual explanations. 
The empirical evaluation of the proposed attack on black-box models trained on real-world datasets demonstrates that they can achieve high-fidelity and high-accuracy extraction even under low query budgets. 
\end{abstract}

%% file: introduction.tex
\section{Introduction}

In recent years, machine learning (ML) models have become prevalent in high stake decision-making systems (\emph{e.g.}, credit scoring~\cite{siddiqi2012credit}, predictive justice~\cite{kleinberg2017human} and hiring~\cite{miller_2015}). 
However, their use is not without any risks as shown by their proven track record of incorrect decisions having consequential impacts on human lives (\emph{e.g.}, people being wrongly denied parole~\cite{wexler2017computer}). 
To address these risks, we have witnessed in the last years an explosion of guidelines -- coming from civil society organizations, the academic world and private companies -- for the ethical development of machine learning~\cite{floridi2019unified,jobin2019global}. 
For instance, to ensure transparency in algorithmic decision processes, the General Data Protection Regulation (GDPR) has an explicit provision requiring explanations of the rationale behind decisions of automatic decision-making systems (among which machine learning is often the key part as mentioned previously) that have a significant impact on individuals~\cite{goodman2016european}. 

Current techniques to achieve transparency include transparent box design and post-hoc explanation of black-box models~\cite{lipton2016mythos,lepri2017fair,montavon2018methods,guidotti2019survey,arrieta2019explainable}. 
Transparent box design aims at building transparent models, which are inherently interpretable~\cite{li2002mining,angelino2017learning,breiman2017classification,ustun2016supersparse}. 
Examples of such models include rules sets~\cite{rijnbeek2010finding,mccormick2011hierarchical,dash2018boolean,li2002mining}, rule lists~\cite{angelino2017learning,yang2017scalable,wang2015falling,aivodji2019learning}, decision trees~\cite{breiman2017classification,narodytska2018learning} and scoring systems~\cite{zeng2017interpretable,ustun2016supersparse,koh2006two}. 

In contrast, post-hoc explanation techniques concern methods used to explain how black-box models produce their outcomes~\cite{guidotti2019survey,arrieta2019explainable}. 
Current families of post-hoc explanations include \emph{global explanations}, \emph{local explanations}, \emph{feature relevance explanations}, \emph{visualization-based explanations} and \emph{example-based explanations}. 
In a nutshell, global explanations break down the whole logic of the black-box model by training an interpretable surrogate model maximizing its fidelity to the black-box model. For instance, decision trees can be used to approximate black-box models~\cite{craven1996extracting}.   
Local explanations aim to explain a single instance by approximating the black-box in the neighbourhood of that instance, also using an interpretable model. Examples of such techniques include LIME~\cite{ribeiro2016should} and SHAP~\cite{lundberg2017unified}.
Feature relevance explanations~\cite{cortez2013using,krause2016interacting} refer to a broad set of methods that help in understanding the black-box model through the analysis of inputs' relative feature importance. 
Visualization-based explanations leverage the use of visualizations to describe the black-box model behavior. 
Examples of such techniques include \emph{saliency maps}~\cite{erhan2009visualizing,baehrens2010explain,simonyan2013deep} used to explain neural network on image classification tasks.
Finally, example-based explanations focus on explaining a black-box model by selecting particular data instances to explain either the behavior of the black-box model or the data distribution. 
Examples of this family include \emph{prototypes and criticisms}~\cite{kim2016examples} and \emph{counterfactual explanations}~\cite{wachter2017counterfactual}, which are the form of explanations we considered in this work. 
More precisely given a black-box model and some input instance, counterfactual explanations are perturbed versions of the input instance that will receive a different prediction by the black-box model.

There exist two fundamental threats to the deployment of post-hoc explanation techniques in real-world applications. 
First, they are subject to explanation manipulations that target the trustworthiness of machine learning models. 
Explanation manipulation attacks leverage post-hoc explanations techniques to give the impression that the black-box model exhibits some good behavior (\emph{e.g.}, no discrimination) while it might not be the case~\cite{adebayo2018sanity,rudin2019stop,aivodji2019fairwashing,fukuchi2019pretending,laugel2019dangers,ghorbani2019interpretation,heo2019fooling,dombrowski2019explanations,merrer2019bouncer,lakkaraju2019fool,slack2019can,zhang2020interpretable}. 
For instance, in~\cite{aivodji2019fairwashing}, the authors have described how it is possible to perform \emph{fairwashing} through global and local explanations' manipulations, which is the possibility that post-hoc explanation techniques could be used to provide cover for unfair black-box ML models. 
In particular, they showed that, given an unfair black-box model $\mathcal{B}$, a dishonest model producer could systematically produce an ensemble of high-fidelity interpretable surrogate models that are fairer than $\mathcal{B}$ according to a predefined notion of fairness, and use those models to justify that the black-box model behaves fairly. 

Second, post-hoc explanations are vulnerable to inference attacks that target the privacy of individuals whose records contributed to the training of machine learning models. 
In this type of privacy attack, the adversary leverages explanations provided by a machine learning model to infer private information such as whether or not a particular individual was part of the training data through a membership inference attack~\cite{shokri2019privacy} or the structure and parameters of the model by performing a model extraction attack~\cite{milli2019model}.

In this work, we focus on the second type of threat by demonstrating how an adversary can use counterfactual explanations to conduct powerful model extraction attacks.
More precisely, we make the following contributions:
\begin{itemize}
\item We provide the first study of model extraction attacks that exploit counterfactual explanations.
\item We introduce different adversary models in this context, which differ in terms of (1) the knowledge of the target model's training data distribution, (2) the knowledge of the target model's architecture and (3) the use of the training data by the explanation algorithm.
\item We study the performances of our attack (\emph{i.e.}, accuracy/fidelity of the surrogate model) under these different adversary models.
\item We demonstrate that our attack can achieve high-fidelity and high-accuracy model extraction under a limited query budget. 
\end{itemize}

The outline of the paper is as follows.
First, in Section~\ref{sec_preliminaries}, we review the background notions on machine learning, counterfactual explanations and model extraction attacks. 
Then, we present in Section~\ref{sec_contrib} our method to devise high-fidelity and high-accuracy model extraction attacks by leveraging counterfactual explanations. 
Afterwards, in Section~\ref{sec:experiments} we report on the evaluation of our attack on black-box models trained on real-world datasets before reviewing the related work in Section~\ref{sec:related}. 
Finally, we discuss potential countermeasures as well as the tension between privacy and explainability in Section~\ref{sec:discussions} and conclude the paper in Section~\ref{sec:conclusion}.

%% file: preliminaries.tex
\section{Preliminaries}
\label{sec_preliminaries}

In this section, we introduce the background notions on machine learning, counterfactual explanations and model extraction attacks necessary to comprehend our work.

\subsection{Machine learning}

A machine learning (ML) model can be defined in a generic manner as a parameterized function $\model{} : \inputspace{} \to  \outputspace{}$, in which $\inputspace{}$ is the input (or feature) space, $\outputspace{}$ the output space and $\param{}$ the parameters (or weights) of the model. 
In this work, we focus on classification tasks within the supervised learning context~\cite{murphy2012machine}. 
In a classification task, the ML model's output is a distribution over $|\outputspace{}|$ classes. 
A supervised learning algorithm aims at building a ML model $\model{}$ from a set of labeled data (\emph{i.e.}, in which the class associated with a particular data instance is known a priori), hereinafter referred to as the \emph{training data}. 
More precisely, given a training data $\traindata{}$ consisting of a sample of independent and identically distributed (i.i.d) pairs $(\singleinput{}_i,\singleoutput{}_i) \in \inputspace{} \times \outputspace{}$ and a loss function $\lossfunc{} : \outputspace{} \times \outputspace{} \to \mathbb{R}^+$, which measures how well the learned model fits the training data $\traindata{}$, an accurate ML model is found by solving the following optimization problem:
\begin{align}
    \hat{ \param{}} = \underset{\param \in \paramspace}{\operatorname{argmin}}  \frac{1}{|\traindata{}|} \Sigma_{i=1}^{|\traindata{}|} \lossfunc(\singleoutput_i, \model(\singleinput_i)) + \lambda\reg(\theta),
 \label{eq1}
\end{align}
in which $\reg(\theta)$ is a regularizer that prevents the ML model from overfitting its training data while $\lambda$ controls the strength of the regularization. 
This optimization problem is the generic framework used to train ML models such as \emph{random forests}~\cite{breiman2001random}, decision lists~\cite{rivest1987learning} and deep neural networks~\cite{Goodfellow-et-al-2016}, which we used in this work.
In a deep neural network (DNN), the function $\model{}$ can be structurally viewed as a hierarchical composition of $k$ parametric function $l_i$, for $i=1,\ldots,k$, in which each function $l_i$ corresponds to a layer of neurons~\cite{Goodfellow-et-al-2016}.

\subsection{Counterfactual explanations}

\begin{figure*}[ht]
\centering
\includegraphics[width=\textwidth]{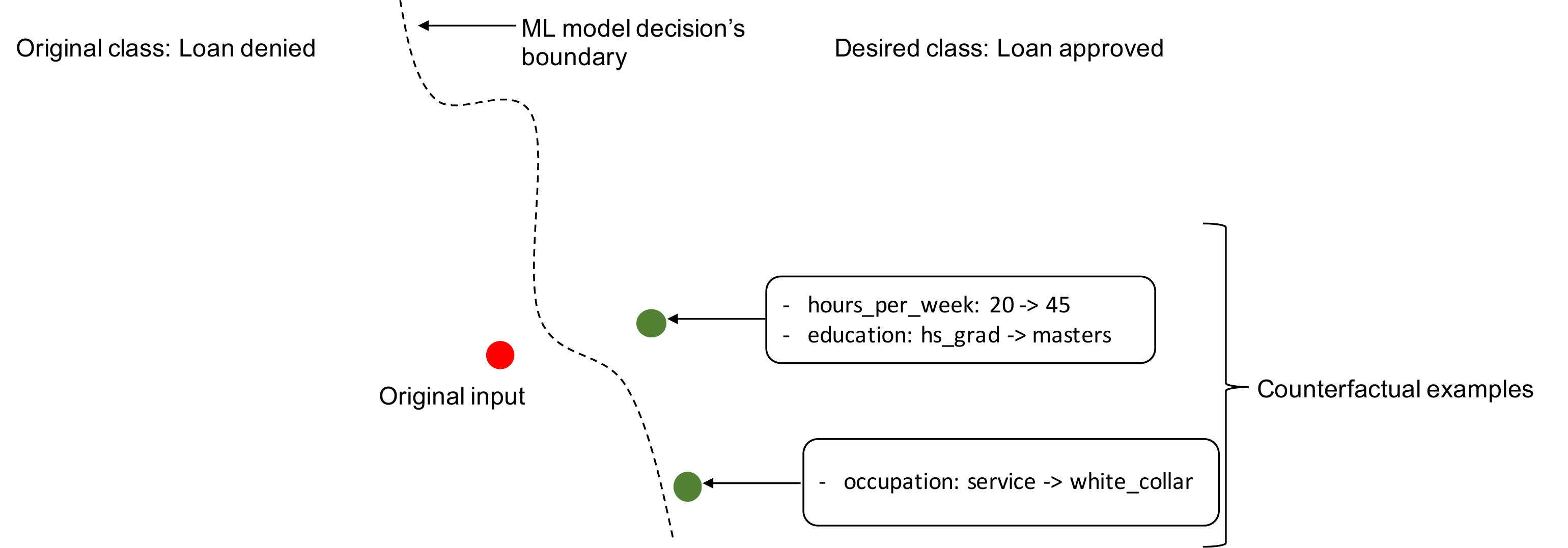}
\caption{Illustration of a counterfactual explanation scenario. 
Given an original instance for which the model predicts the \emph{loan denied} class, a counterfactual explanation framework provides different instances that are close to the original one but belong to the desired class (\emph{loan approved} here). 
An individual asking for an explanation can thus see which aspects of his profile he may try to change to yield the desired outcome.}
\label{fig:cf_illustration}
\end{figure*}

\input{tables/cf_eg}

Counterfactual explanations~\cite{wachter2017counterfactual,lash2017generalized,laugel2017inverse,tolomei2017interpretable,grath2018interpretable,russell2019efficient,ustun2019actionable,joshi2019towards,pawelczyk2020learning,mothilal2020explaining,karimi2020model} 
are data instances that are close to the input instance to be explained but whose model predictions are different from that of the input instance. 
More precisely, given a black-box model $\bbox{}$, an original input $x_0$, its predicted outcome $y_0=\bbox(x_0)$ and a desired outcome $y \neq y_0$, a counterfactual explanation $c(x_0)$ for the input $x_0$ is usually obtained by solving the following optimization problem:
\begin{align}
   c(x_0) = \underset{c}{\operatorname{argmin}}  ~ \lossfunc(\bbox(c),y) + |c - x_0|,
\end{align}
in which $\lossfunc(\bbox(c),y)$ ensures that the obtained counterfactual $c(x_0)$ has a different prediction from that of the original input $x_0$ while the second term $|c - x_0|$ helps in obtaining a counterfactual close to the original instance. 
Figure~\ref{fig:cf_illustration} illustrates a counterfactual explanation scenario while Table~\ref{tab:cf_eg} provides concrete examples of counterfactual explanations obtained on a real world dataset, namely Adult Income~\cite{frank2010uci}.

\textbf{Diverse counterfactuals.} To be more actionable, counterfactual explanation frameworks often generate for each input instance, several counterfactuals covering a diverse range of possibilities instead of the single closest one~\cite{wachter2017counterfactual}. 
Providing diverse counterfactuals allows users to decide the most efficient way by which they can influence their profile to obtain the desired outcome. 
At the same time, on the privacy side, it also leaks more information to the adversary and enables him to mount a more powerful attack. 
In this paper, we rely on the \emph{DiCE} framework~\cite{mothilal2020explaining} to implement the explanation API of the target models. 
Nonetheless, the proposed attack is generic enough to work with any counterfactual explanation framework.

In a nutshell, DiCE aims to find valid and actionable counterfactual examples by solving the following optimization problem:
\begin{eqnarray}
   C(x_0) = &\underset{c_1,\ldots,c_k}{\operatorname{argmin}}  ~ \frac{1}{k} \Sigma_{i=1}^k \lossfunc(\bbox(c_i),y) + \frac{\lambda_1}{k} \Sigma_{i=1}^k|c_i - x_0| \nonumber \\ & -\lambda_2 \operatorname{dpp\_diversity}(c_1,\ldots,c_k),
\end{eqnarray}
in which $\mathcal{B}$ is the black-box model, $x_0$ is the original input to be explained, $y \neq \mathcal{B}(x_0)$ is the desired outcome, $c_i$ is a counterfactual example and $k$ is the number of counterfactuals to return.
The loss function $\lossfunc(\bbox(c_i),y)$ ensures that each of the counterfactuals has a different outcome than that of the original input $x_0$ while $|c_i - x_0|$ leads to the counterfactual being close to the original input. 
Finally, $\operatorname{dpp\_diversity}(\cdot)$ is the diversity metric while $\lambda_1\in \mathbb{R}^+$ and $\lambda_2 \in \mathbb{R}^+$ are the hyperparameters used to balance the proximity and diversity. 
More precisely, the larger $\lambda_1$ is, the closer the counterfactuals will be to to the query instance.
Similarly, the larger $\lambda_2$ is, the more diverse the counterfactuals return will be diverse.

In this paper, we will investigate the success rate of the attack we propose in both the cases of single and diverse counterfactuals. 

\subsection{Model extraction}
\label{subsec:modelextraction}

A model extraction attack is an inference attack in which an adversary $\adv{}$ obtains a surrogate model $\surr{}_\adv{}$ that is similar to the targeted model $\bbox{}$. 
The precise meaning of the \emph{similarity} depends on the adversary's objective, while the success of the attack depends on the adversary's capabilities.

\textbf{Adversary objective.} Previous works~\cite{atli2019extraction,jagielski2020high} have considered two main categories of model extraction attacks depending on the goal of the adversary, namely \emph{accuracy-based} and \emph{fidelity-based} model extraction attacks. 
In accuracy-based model extraction attacks, also known a \emph{theft-motivated model extraction} attacks~\cite{jagielski2020high}, the adversary aims at learning a surrogate model $\surr{}_\adv{}$ whose accuracy is as close as possible to that of the target's model $\bbox{}$. 
Typically here, model extraction provides a financial benefit to the adversary as he can use the surrogate model as a substitute for the commercial API of his target. 
In fidelity-based model extraction attacks, also known as \emph{reconnaissance-motivated model extraction} attacks~\cite{jagielski2020high}, the objective of the adversary is to build a surrogate model $\surr{}_\adv{}$ maximizing the \emph{fidelity} with the target's model $\bbox{}$. 
The fidelity $\operatorname{Fid}(\surr{}_\adv{})$ of the surrogate is defined as its accuracy relative to $\bbox{}$ over a reference set $X_r \subset \inputspace{}$~\cite{craven1996extracting}:
\begin{align}
\operatorname{Fid}(\surr{}_\adv{}) = \frac{1}{|X_r|} \sum_{x \in X_r} \mathbb{I}(\surr{}_\adv{}(x)=\bbox{}(x)).
\end{align}
In this context, a model extraction attack is often a first step towards mounting other attacks such as a model inversion attacks~\cite{fredrikson2014privacy,fredrikson2015model} or adversarial examples discovery~\cite{szegedy2013intriguing,goodfellow2014explaining,papernot2017practical}.

A particular case of fidelity-based model extraction attack, known as \emph{functionally equivalent extraction}, occurs when the adversary is able to build a surrogate $\surr{}_\adv{}$ matching the predictions of the target's model $\bbox{}$ over the whole input space (\emph{i.e.}, $\forall x \in \inputspace{}, \surr{}_\adv{}(x)=\bbox{}(x)$). 
As pointed out in~\cite{jagielski2020high}, functionally equivalent extraction attacks require model-specific techniques. 
In contrast, both accuracy-based and fidelity-based model extraction attacks generally rely on the flexibility of learning-based approaches, making them more generic. 
In the latter case, the target's model $\bbox{}$ is used as a labeling oracle by the adversary.

\textbf{Adversary capabilities.} Following the taxonomy introduced in~\cite{jagielski2020high}, we describe the adversary capabilities around three axes, namely the \emph{domain knowledge}, the \emph{deployment knowledge}, and the \emph{model access}. 
Domain knowledge corresponds to the adversary's prior information on the task of the target model. 
For learning-based approaches, a common assumption is that the adversary knows as much about the task as the designer of the target model.
Deployment knowledge refers to the adversary's knowledge of the target model's characteristics (\emph{e.g.}, architecture, training dataset, training algorithm, hyperparameters, \ldots).  
Finally, the model access indicates how the adversary interacts with the target's model and the form of information extracted from these interactions. 
More precisely, this includes both the number of queries the adversary is allowed to make to the target's model and the type of the model's output (\emph{e.g.}, labels, probabilities, gradients, counterfactual explanations,\ldots). 

%% file: tables/cf_eg.tex
\begin{table*}[ht]

\resizebox{\textwidth}{!}{%
\begin{tabular}{@{}c|ccccccccccc@{}}
\cmidrule(l){2-12}
 & \textbf{Age} & \textbf{Workclass} & \textbf{Education} & \textbf{Marital status} & \textbf{Relationship} & \textbf{Occupation} & \textbf{Race} & \textbf{Gender} & \textbf{Capital gain} & \textbf{Capital loss} & \textbf{Hours per week} \\ \midrule
\begin{tabular}[c]{@{}c@{}}Original input\\ (outcome: $\leq 50$K)\end{tabular} & $33$ & Private & Assoc-acdm & Married & Own-child & Professional & White & Female & $0$ & $0$ & $40$ \\ \midrule
\multirow{2}{*}{\begin{tabular}[c]{@{}c@{}}Counterfactuals\\ (outcome: $>50$K)\end{tabular}} & - & - & Doctorate & - & - & - & - & - & $33703$ & - & $39$ \\
 & - & - & - & - & - & White-collar & - & - & $99985$ & $4333$ & - \\ \bottomrule
\end{tabular}%
}

\caption{\label{tab:cf_eg} Examples of counterfactuals obtained on Adult Income~\cite{frank2010uci} dataset. 
The task is to predict whether an individual earns more than 50,000\$ per year. 
The top row corresponds to the different features of the input instance. The second row depicts the data instance to be explained as well as its original outcome. 
Finally, the last two rows are examples of counterfactuals generated to explain the original input. 
Dashed marks refer to features that are unchanged.}
\end{table*}

%% file: contributions.tex
\section{Model extraction from counterfactual explanations}
\label{sec_contrib}

In this section, we first frame the generic problem of explanation-based model extraction before presenting the particular case of counterfactual explanation, which is the focus of this work.
Afterwards, we describe the different adversarial models investigated in our work before describing their corresponding model extraction attacks. 

\subsection{Problem formulation}

\begin{figure*}[ht]
\centering
\includegraphics[width=\textwidth]{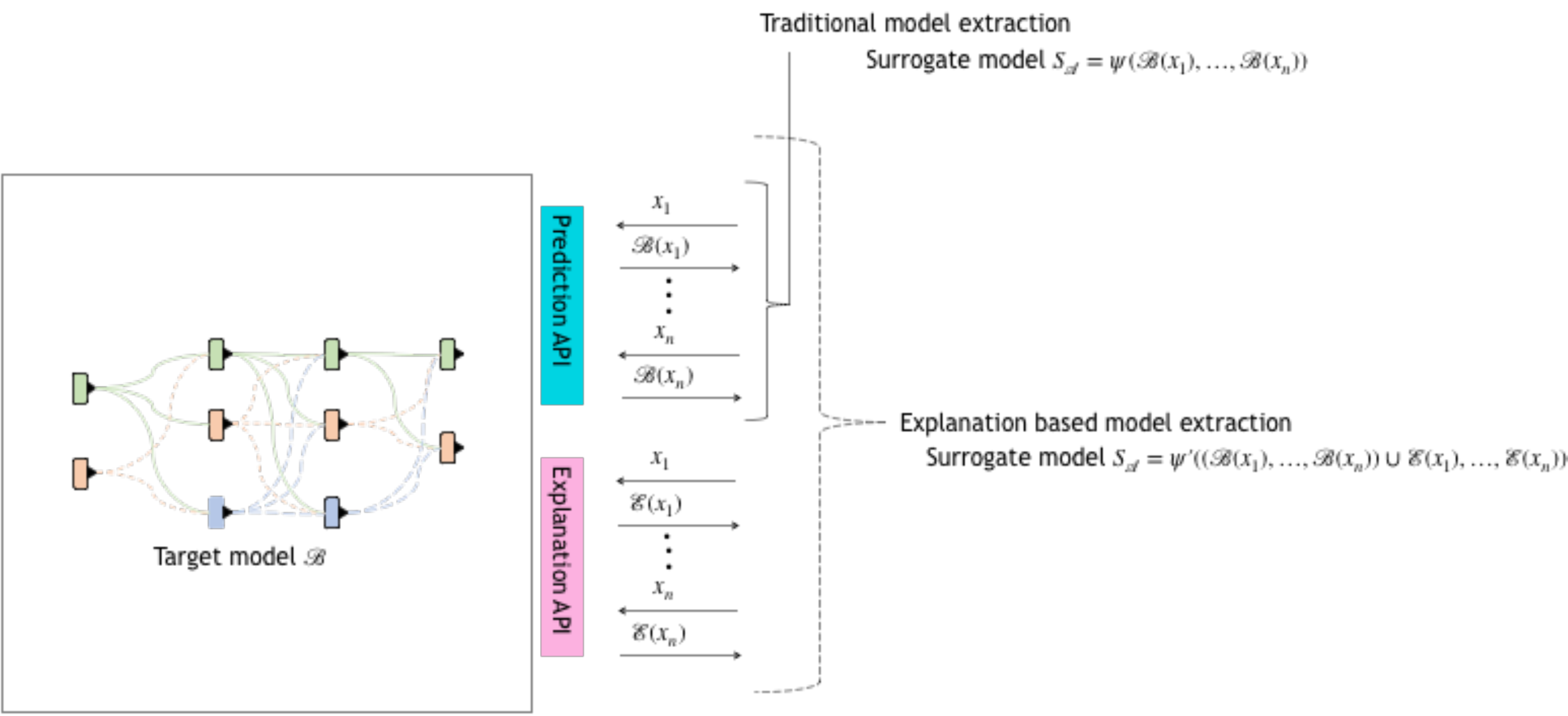}
\caption{Illustration of a traditional model extraction attack and an explanation-based model extraction. 
In the former, the adversary relies on the predictions $\mathcal{B}(x_1),\ldots,\mathcal{B}(x_n)$ of the target model $\mathcal{B}$ to build the surrogate model $S_\adv{}$ using a process $\psi(\cdot)$, while in the later, the adversary combines the predictions $\mathcal{B}(x_1),\ldots,\mathcal{B}(x_n)$ and the explanations $\mathcal{E}(x_1),\ldots,\mathcal{E}(x_n)$ of the target model $\mathcal{B}$ to generate the surrogate $S_\adv{}$ using another process $\psi'(\cdot)$.}

\label{fig:problem_form}
\end{figure*}

\begin{figure*}[ht]
\begin{subfigure}[t]{.22\textwidth}
        \centering
        \includegraphics[width=\linewidth]{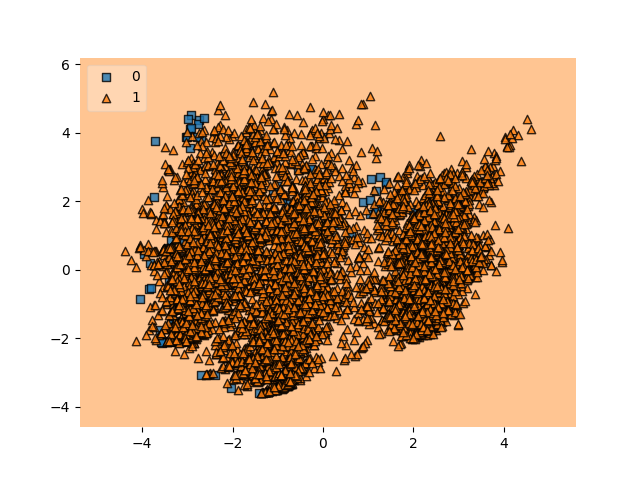}
        \caption{Tradition model extraction with $8059$ queries}
        \label{subfig:baseline_boundary}
\end{subfigure}\hfill
\begin{subfigure}[t]{.22\textwidth}
        \centering
        \includegraphics[width=\linewidth]{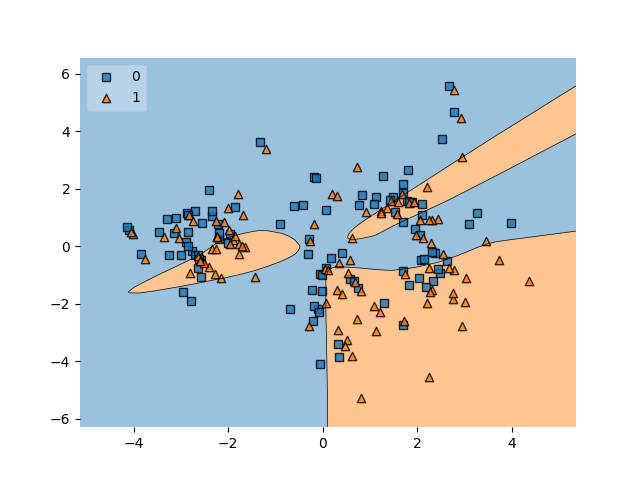}
        \caption{Counterfactual-based model extraction with $100$ queries.}
        \label{subfig:cf_100_boundary}
\end{subfigure}\hfill
\begin{subfigure}[t]{.22\textwidth}
        \centering
        \includegraphics[width=\linewidth]{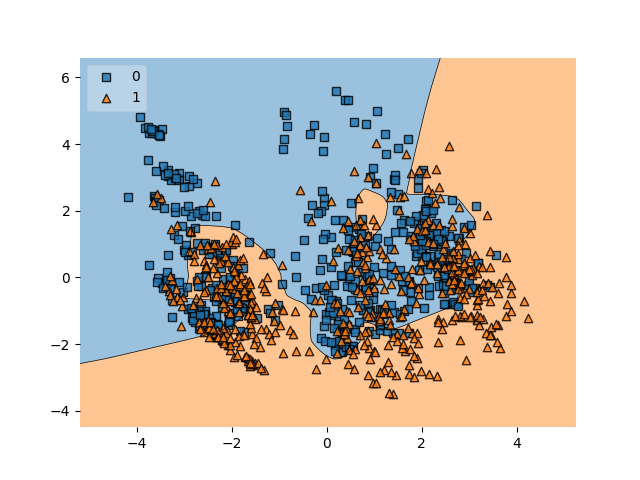}
        \caption{Counterfactual-based model extraction with $500$ queries.}
        \label{subfig:cf_500_boundary}
\end{subfigure}\hfill
\begin{subfigure}[t]{.22\textwidth}
        \centering
        \includegraphics[width=\linewidth]{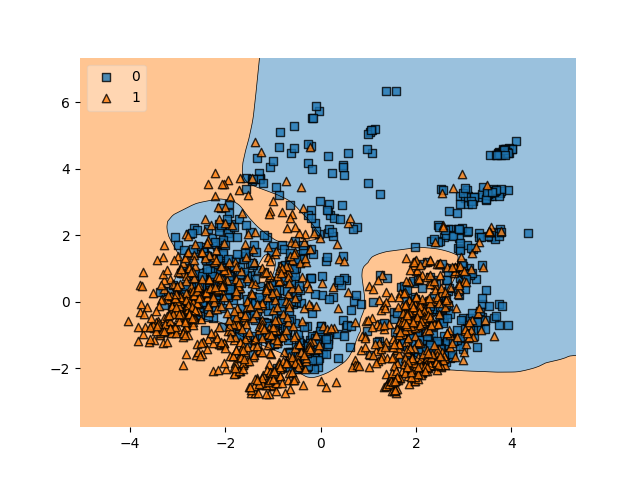}
        \caption{Counterfactual-based model extraction with $1000$ queries.}
        \label{subfig:cf_1000_boundary}
\end{subfigure}%
\caption{Decision boundary of the surrogate model on Adult Income dataset~\cite{frank2010uci}.}
\label{fig:decision_boundaries}
\end{figure*}

As illustrated in Figure~\ref{fig:problem_form}, in an explanation-based model extraction attack, the adversary leverages both the predictions and the explanations of the target model to build the surrogate model.

\begin{definition}[Explanation-based model extraction]
\label{def:expl-extraction}
Given a target model $\mathcal{B}$, its prediction API $\mathcal{B}(\cdot)$ as well as its explanation API $\mathcal{E}(\cdot)$, both available in a black-box setting, a set of data points $x_1,\ldots,x_n$, the explanation-based extraction attack consists in using both the explanations and the predictions of the target model to build a surrogate $S_\adv{} \approx \mathcal{B}$, using an attack process $\psi(\cdot)$.
\end{definition}

In the particular context of counterfactual explanations, the explanation API $\mathcal{E}(\cdot)$ returns for each data point $x_i$ its corresponding counterfactual explanation $c(x_i)$ along with its associated outcome $\overline{y_i}$. 
In the case of diverse counterfactuals, the explanation API will return a set $C(x_i)$ of counterfactual examples instead of a single one.

\subsection{Attack description}
\label{subsec:attack_description}

\textbf{Adversary model.} We are interested in a fidelity-based extraction attack (also called reconnaissance-motivated extraction attack) in which the adversary $\adv{}$ will rely on both the predictions and the counterfactual explanations of the target model to conduct his attack. 
Similarly to \cite{jagielski2020high}, we will assume that the adversary knows as much about the task as the designer of the target model in terms of domain knowledge.
As for the model access, the adversary will have black-box access to the target model's predictions and counterfactual explanations. 
We also assume a bound on the number of queries that the adversary is allowed to make. 
Each query to the explanation API $\mathcal{E}(\cdot)$ returns one or more counterfactual explanations depending on the diversity criteria.
Finally, for the deployment knowledge, we consider different scenarios according to (1) the knowledge of the training data distribution, which can be \emph{known}, \emph{partially known} (\emph{e.g.}, knowledge of the marginal distribution) or \emph{unknown}, (2) the knowledge of the target model architecture (\emph{known} or \emph{unknown}), and (3) the use of the training data by the explanation algorithm (\emph{used} or \emph{unused}).

\textbf{Attack strategy.} To conduct his attack, the adversary first builds his \emph{attack set} $D_\adv{}$ according to his knowledge of the distribution of the target model's training data. 
Then, for each data point $x \in D_\adv{}$, he sends a query to both the prediction API $\mathcal{B}(\cdot)$ and the explanation API $\mathcal{E}(\cdot)$ of the target model. 
Finally, $\adv{}$ trains the surrogate model $S_\adv{}$ according to his knowledge of the target model's architecture, by using a \emph{transfer set} $\mathcal{T}_\adv{} = \{D_\adv{}, \mathcal{B}(D_\adv{})\} \cup \mathcal{E}(D_\adv{})$ consisting of both the outputs of the prediction and explanation APIs.

In traditional model extraction attacks, the transfer set $\mathcal{T}_\adv{}$ of the adversary can be imbalanced due to the unequal distribution of classes within the dataset. 
As a result, there may be a significant difference between the class-based accuracy of the surrogate model $S_\adv{}$ and the target model $\mathcal{B}$~\cite{atli2019extraction}. 
In contrast, counterfactual explanations-based model extractions attack do not suffer from such limitations as the attack set is balanced by construction since each instance is followed by its corresponding counterfactual explanation. 
As illustrated in Figure~\ref{fig:decision_boundaries}, the surrogate models of the adversary can better approximate the decision boundary of the target model with few queries in the counterfactual-based model extraction attacks compared to traditional model extraction ones.

%% file: experiments.tex
\section{Experimental evaluation}
\label{sec:experiments}

In this section, we report on the performances of counterfactual explanations-based model extraction attacks when evaluated on real datasets.

\subsection{Experimental setting}

\textbf{Datasets.} We have conducted our experiments on three public datasets that are extensively used in the \emph{FaccT} (Fairness, Accountability, and Transparency) literature, namely \emph{Adult Income}~\cite{frank2010uci}, \emph{COMPAS}~\cite{angwin2016machine} and \emph{Default Credit}~\cite{frank2010uci}. 
In a nutshell, the Adult Income dataset contains information about individuals collected from the 1994 U.S. census. 
The dataset contains 48,842 individuals, each described by 11 attributes. 
The underlying classification task is to predict whether or not an individual makes more than 50,000$\$$ per year in terms of income. 
The COMPAS dataset gathers records from criminal offenders in Florida during 2013 and 2014. 
The dataset contains 7,214 individuals, each described by 8 attributes. 
The classification task considered is to predict whether a subject will re-offend within two years after being released. 
Finally, the Default Credit dataset is composed of information on Taiwanese credit card users. 
The dataset contains 29,986 individuals, each described by 23 attributes, while the classification task is to predict whether a user will default in his payments. 

\textbf{Evaluation metrics.} Our main objective is to conduct a reconnaissance-motivated model extraction attack. As such, we will use the fidelity metric, as described in Section~\ref{subsec:modelextraction}, as our primary evaluation metric for the success of the attack. 
Nonetheless, we will also report on the accuracy of the surrogate.

\textbf{Black-box models.} Each dataset is split into three subsets, namely the \emph{training sets} ($67\%$), the \emph{testing sets} ($16.5\%$), and the \emph{attack pools} ($16.5\%$). 
The black-box models are learned on the training sets. 
The testing sets are used to evaluate (1) the accuracy of both black-box models and surrogates models and (2) the fidelity of the surrogate model relative to the target black-box model. 
The attack pools are used only for the scenario in which the adversary is assumed to know the data distribution. 
For both Adult Income and COMPAS, the target models are Multi-Layer Perceptrons (MLPs) with two hidden layers, with respectively 75 and 50 neurons. 
For Default Credit, the target model is a MLP with one hidden layer of 50  neurons. 

For all the three target models, we have used the L1 regularization (with $\lambda=0.001$), the RMSprop optimizer~\cite{tieleman2012lecture}, the rectifier activation function (\emph{ReLu}) for hidden layers, the \emph{Sigmoid} activation function for output layers and train the models for 100 epochs. 
Table~\ref{tab:bbox_summary} summarizes the accuracy of the three black-box models on their training and test sets.

\input{tables/bbox_summary}

\textbf{Scenarios investigated.} 
The adversary model presented in Section~\ref{subsec:attack_description} leads us to consider five different counterfactual-based model extraction scenarios, namely (S1) single counterfactual with known training data distribution, (S2) single counterfactual with partially known training data distribution, (S3) single counterfactual with unknown training data distribution, (S4) multiple counterfactuals with known training data distribution and (S5) impact of the proximity and diversity metrics on the performances of the model extraction.
The first three scenarios are variants of the same setting in which the explanation API only provides a single counterfactual explanation per query, but under different assumptions on the adversary knowledge on the distribution of the training data of the target model. 
The objective of the last two scenarios is to study the impact on the success rate of the extraction attack of having access to multiple and diverse counterfactual explanations per query. 

For all five scenarios, the performances are evaluated according to the adversary's knowledge on the architecture of the target model and whether or not the explanation API uses the training data. 
When the adversary does not know the target model's architecture, we imagine that typically the adversary will have a trial-and-error strategy in which different architectures will be tried with the one maximizing the fidelity of the surrogate being kept at the end. 
In our experiments, we simulate this situation with an adversary that tries $5$ different architectures, which we describe in Table~\ref{tab:arch}. 
Remark that since the surrogate training is done offline once the transfer set has been built, the adversary is only limited in its exploration by its computational resources and the time he is willing to dedicate to this exploration.
In particular, if he has the sufficient resources, he might even use advanced techniques for exploring the space of possible architectures such as \emph{Neural Architecture Search}~\cite{elsken2019neural} to maximize the fidelity of the surrogate model.

Hereafter, we detail each of the five scenarios.

\textbf{(S1) Single counterfactual with known training data distribution.} 
In this scenario, the adversary directly uses the attack pool as his attack set $D_\adv{}$. 
More precisely, he selects a subset $Q_\adv{}$ of $D_\adv{}$ to query the target model and construct his transfer set $\mathcal{T}_\adv{} = \{Q_\adv{}, \mathcal{B}(Q_\adv{})\} \cup \mathcal{E}(Q_\adv{})$. 
In the experiments conducted, we have considered different values $|Q_\adv{}| \in \{100, 250, 500, 1000\}$ for the number of queries to study its effect on the attack's performance. 
For each value of $|Q_\adv{}|$, the experiment is repeated over $10$ random sampling of $Q_\adv{}$ and the average fidelity and accuracy of the surrogate are reported. 
Additionally, we compare the performances of the surrogate with a baseline model trained using the complete attack pool $D_\adv{}$ and the predictions $\mathcal{B}(D_\adv{})$ of the target model.

\textbf{(S2) Single counterfactual with partially known training data distribution.} 
Here, the adversary is assumed to know the marginal distribution of the attributes of the training set. 
To perform his attack, in this scenario, the adversary builds an attack set $D_\adv{}$, composed of data points sampled according to the marginal distribution of the attributes. 
The rest of the attack is similar to the process described above for (S1).

\textbf{(S3) Single counterfactual with unknown training data distribution.} 
This scenario is similar to (S2) except that the distribution of the training data of the target model is unknown. 
As a consequence, the attack set $D_\adv{}$ is generated simply by uniformly sampling data points from the input space.
Clearly, this can sometimes lead to the generation of unrealistic data points.

\textbf{(S4) Multiple counterfactuals with known training data distribution.} 
In this scenario, the same configuration used in (S1) is considered, but the number $k$ of counterfactuals provided by the explanation API is increased. 
More precisely, the attack performances are studied for $k$ in the range $\{3, 5, 7\}$. 
For each of these settings, the default values for the proximity and diversity hyperparameters are used (\emph{i.e.}, $\lambda_1=0.5$ and $\lambda_2=1.0$). 

\textbf{(S5) Impact of the proximity and diversity on the performances of the model extraction.} 
In this scenario, the impact of proximity and diversity on the surrogate model's performance is explored. 
For the sake of simplicity, we focus on the setting in which the adversary knows the data distribution and the training data is used by the explanation API since the results are similar in both cases. 
We set $|Q_\adv{}|=1000$, $k=5$, $\lambda_1 \in \{0.5, 1.0, 1.5, 2.0, 2.5, 3.0\}$ and $\lambda_2 \in \{1.0, 1.5, 2.0, 2.5, 3.0\}$.

All our experiments were run on an Intel Core i7 (2.90 GHz, 16GB of RAM) laptop. We provide a software implementation to reproduce the results of our experiments at \href{https://github.com/aivodji/mrce}{https://github.com/aivodji/mrce}.

\input{tables/arch}

\subsection{Experimental results}
\input{tables/s1}

\input{tables/s2}

\input{tables/s3}

\begin{figure*}[ht]
\begin{subfigure}[t]{.22\textwidth}
        \centering
        \includegraphics[width=\linewidth]{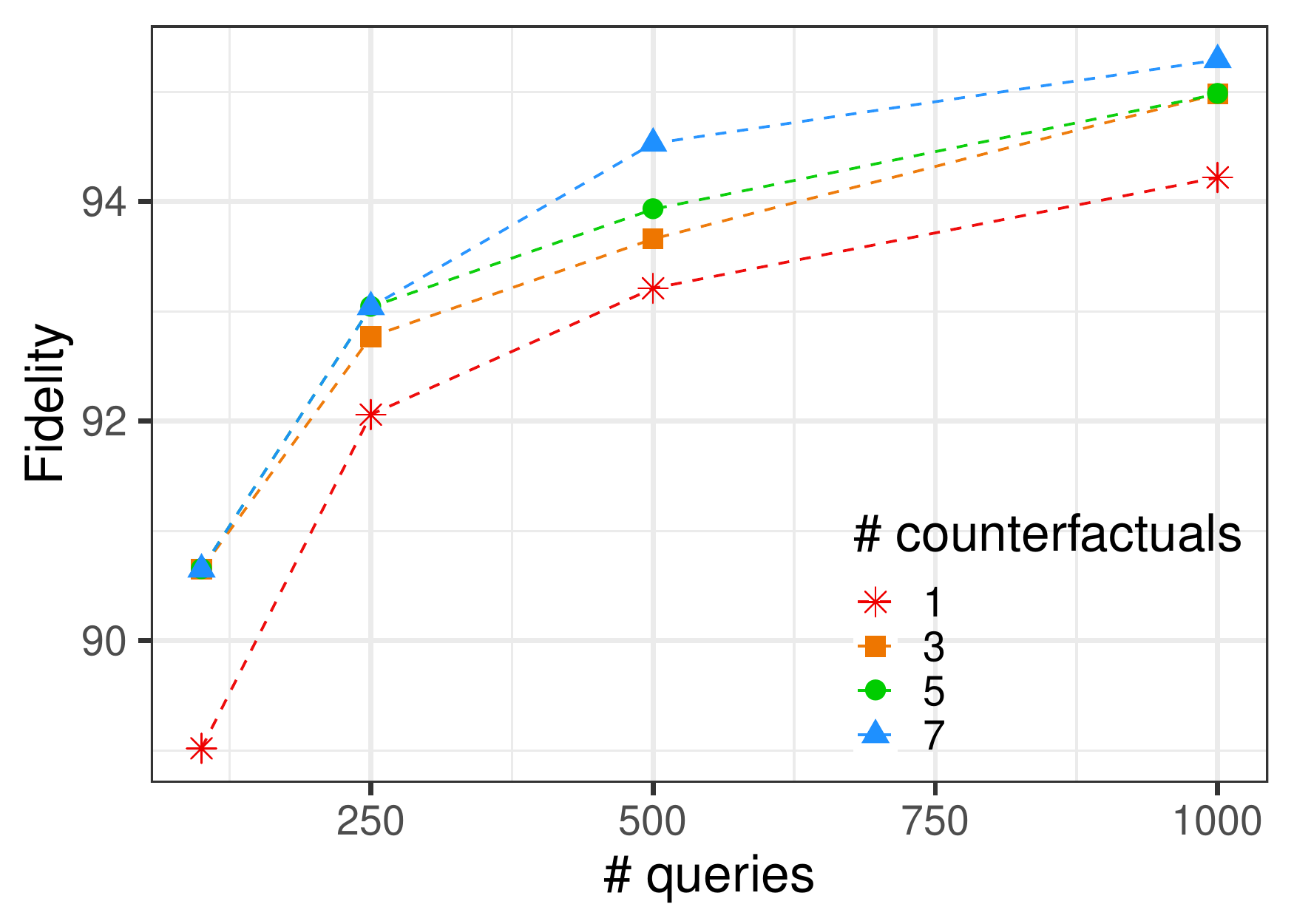}
        \caption{Arch. known, training data used by $\mathcal{E}(\cdot)$}
        \label{subfig:known_public}
\end{subfigure}\hfill
\begin{subfigure}[t]{.22\textwidth}
        \centering
        \includegraphics[width=\linewidth]{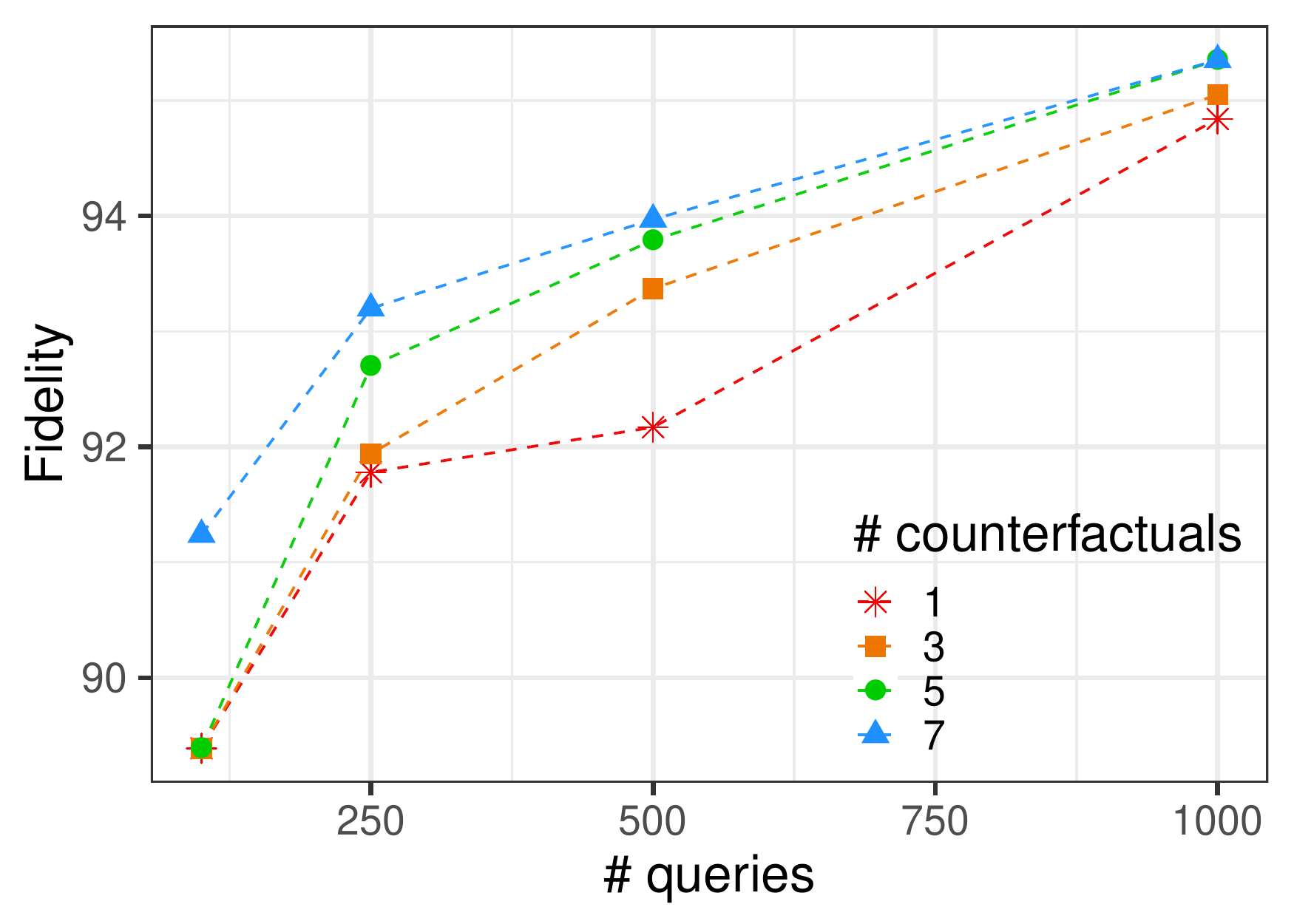}
        \caption{Arch. known, training data not used by $\mathcal{E}(\cdot)$}
        \label{subfig:known_private}
\end{subfigure}\hfill
\begin{subfigure}[t]{.22\textwidth}
        \centering
        \includegraphics[width=\linewidth]{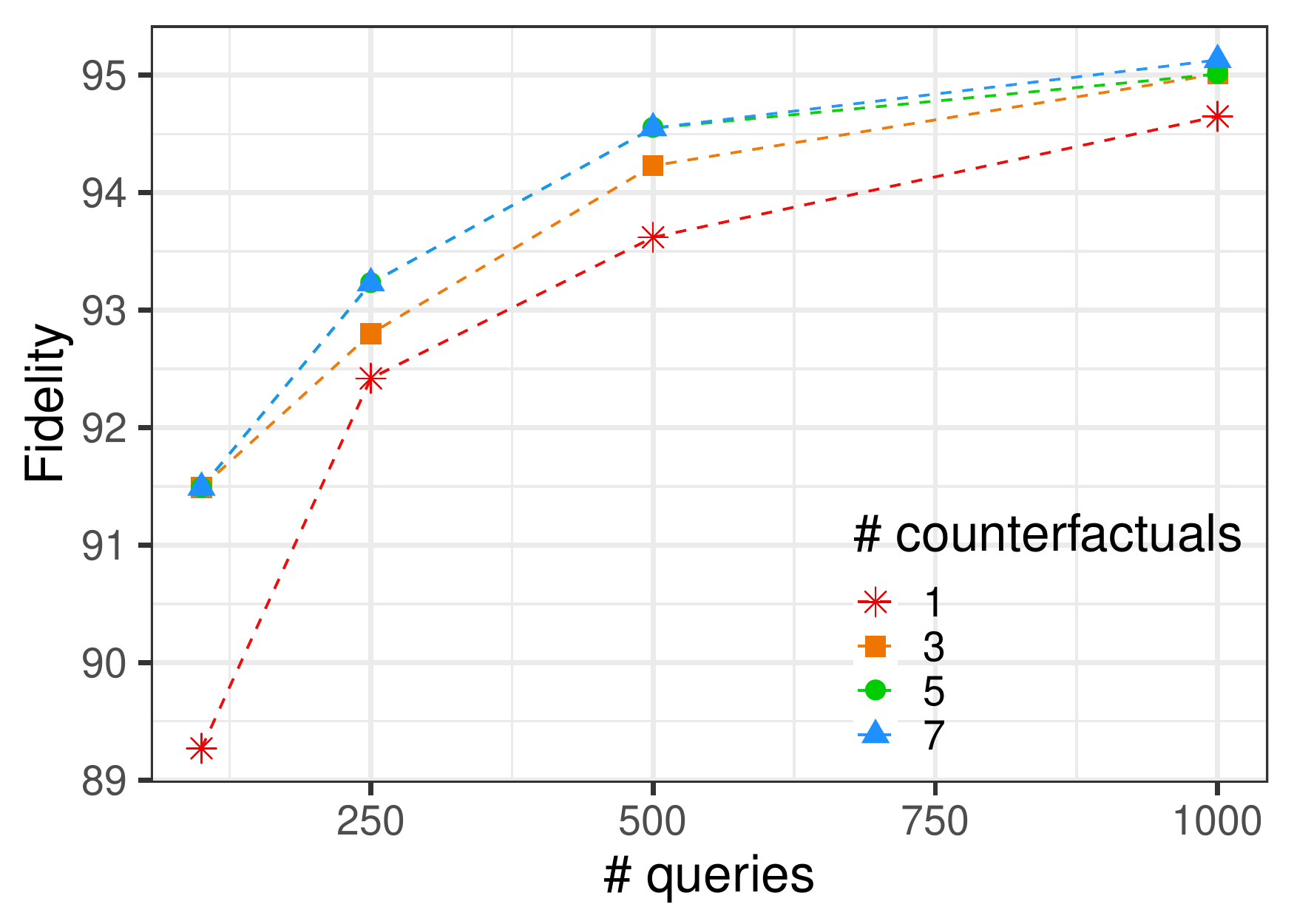}
        \caption{Arch. unknown, training data used by $\mathcal{E}(\cdot)$}
        \label{subfig:unknown_public}
\end{subfigure}\hfill
\begin{subfigure}[t]{.22\textwidth}
        \centering
        \includegraphics[width=\linewidth]{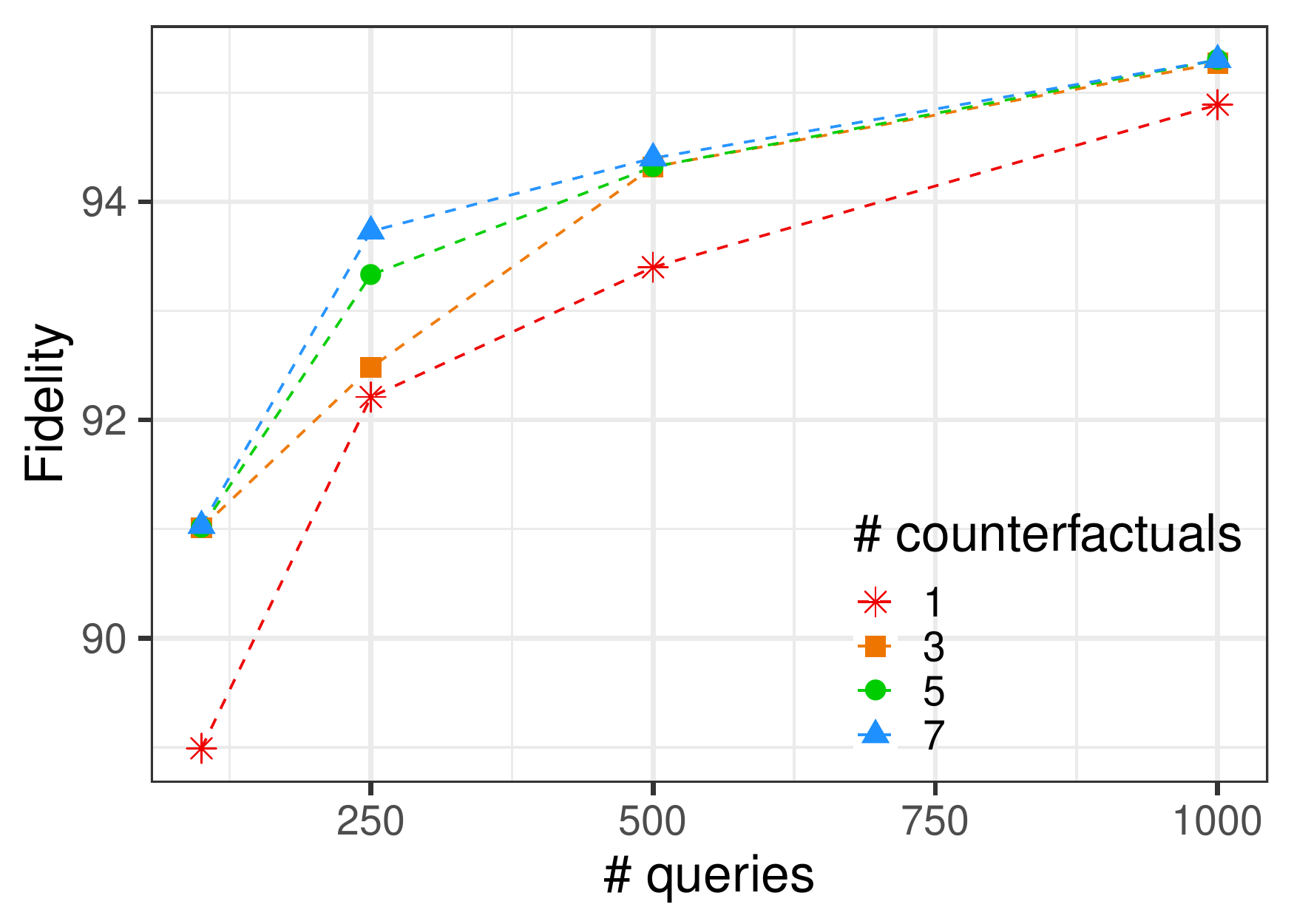}
        \caption{Arch unknown, training data not used by $\mathcal{E}(\cdot)$}
        \label{subfig:unknown_private}
\end{subfigure}%
\caption{Performances (\emph{i.e.}, fidelity) of the model extraction attack in scenario (S4) for Adult Income. 
Results demonstrate the impact of the number of counterfactual explanations per query on the extraction attack's fidelity.}
\label{fig:ncfs_impact}
\end{figure*}

\begin{figure*}[h!]
\begin{subfigure}[t]{.5\textwidth}
        \centering
        \includegraphics[width=\linewidth]{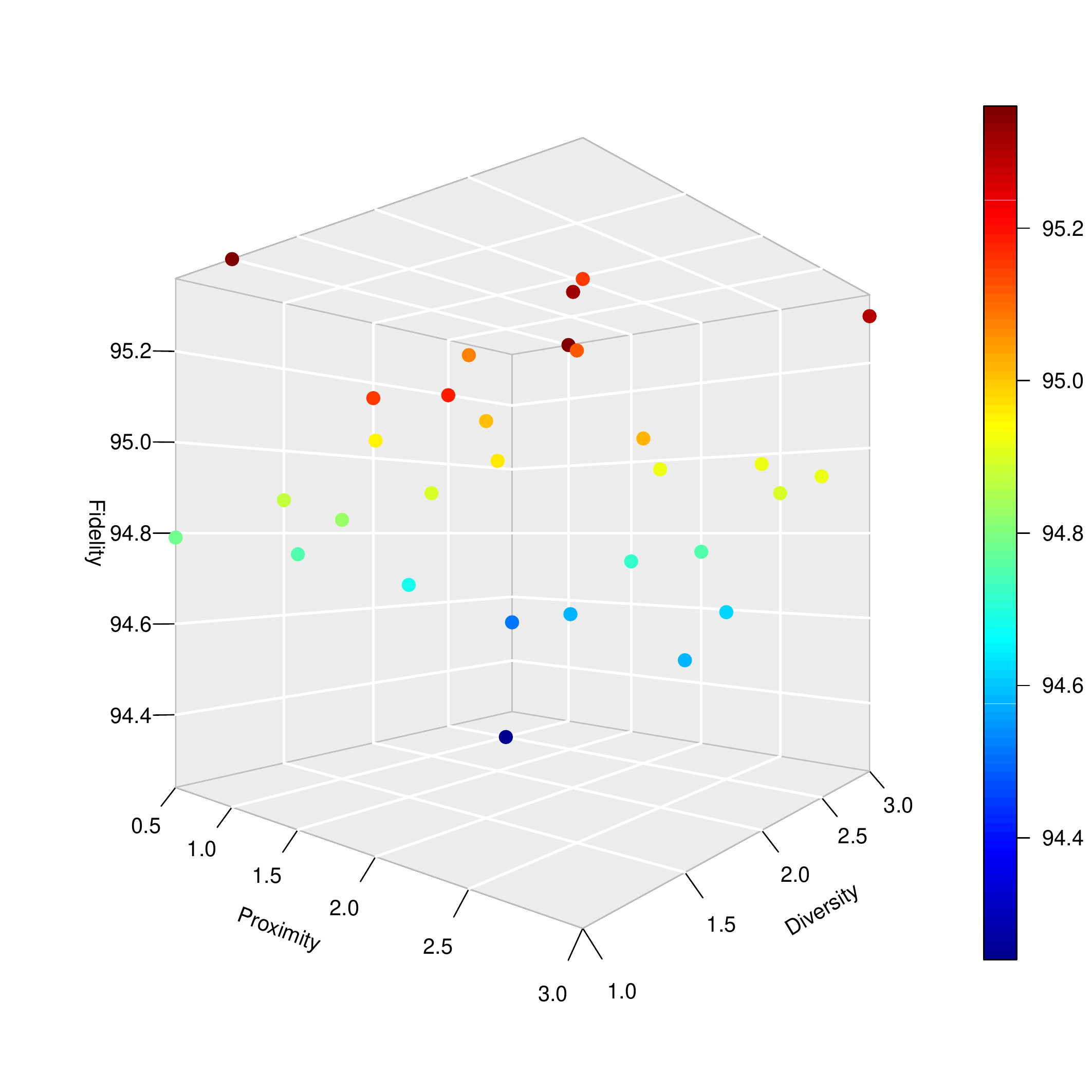}
        \caption{Architecture known}
        \label{subfig:cf_known}
\end{subfigure}\hfill
\begin{subfigure}[t]{.5\textwidth}
        \centering
        \includegraphics[width=\linewidth]{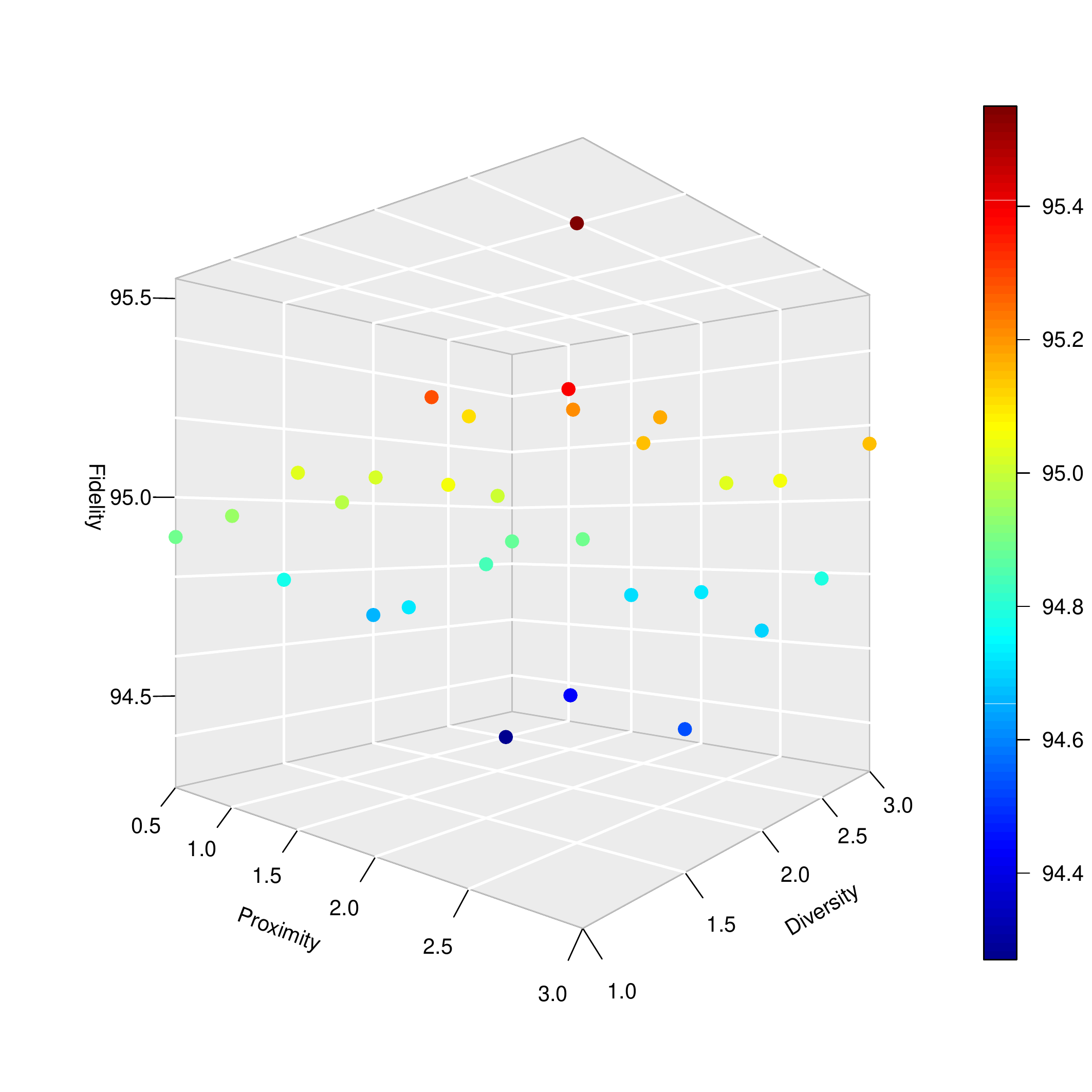}
        \caption{Architecture unknown}
        \label{subfig:cf_unknown}
\end{subfigure}%
\caption{Performances (fidelity) of the model extraction attack in scenario (S5) for Adult Income dataset. 
The results show the impact of the proximity and the diversity metrics on the fidelity of the surrogate.}
\label{fig:diverse_cf}
\end{figure*}

\textbf{(S1) Single counterfactual with known training data distribution.} 
Table~\ref{tab:s1} summarizes the results obtained for scenario (S1). 
The attack is evaluated on Adult Income, COMPAS and Default Credit datasets. 
Overall, for all these three datasets, we observe that with only $250$ queries, our attack reaches a fidelity of $90\%$. 
This fidelity is higher than that of the baseline, which is a traditional model extraction attack with $8059$ queries for Adult Income, $1192$ queries for COMPAS and $4948$ queries for Default Credit. 
We also observed that as the number of queries increases, both the fidelity and the accuracy of the surrogate also improve.
With only $1000$ queries, our attack already reaches a fidelity of $94\%$ on Adult Income, $93\%$ on COMPAS and $98\%$ on Default Credit and an accuracy matching that of the target model (as measured on its test set) on all three datasets.  
Moreover, an interesting finding of our study is that the knowledge of the target model architecture and the use of the training data by the explanation API does not lead to a significant advantage with respect to the attack's success.

\textbf{(S2) Single counterfactual with partially known training data distribution.} 
Table~\ref{tab:s2} displays the results obtained for scenario (S2). 
Here, for the sake of simplicity, we have only performed the experiments on the Adult Income dataset. 
The results demonstrate that an adversary who only knows the features' marginal distribution can still perform a powerful model extraction attack. 
In particular, with $1000$ queries, the surrogate model $S_\adv{}$ still reaches a fidelity of $93\%$ and an accuracy close to that of the target model on the test set.  

\textbf{(S3) Single counterfactual with unknown training data distribution.} 
Table~\ref{tab:s3} describes the performance of our attack for scenario (S3). 
Similarly to (S2), we focus on the Adult Income dataset. 
Overall, the results show that even without knowing the data. distribution, the adversary can build a surrogate model performing better than the one obtained using a traditional extraction attack with $8\times$ more labels and with full knowledge of the data distribution. 
However, compared to the fidelity of counterfactual-based extraction attacks with partial knowledge (respectively full knowledge) of the data distribution, the surrogate's fidelity decreases by $7.79\%$ (respectively $9.14\%$).

\textbf{(S4) Multiple counterfactuals with known training data distribution.} 
Figure~\ref{fig:ncfs_impact} describes the impact of the number of counterfactuals provided for each query on the performance of the extraction attack. 
Overall, we can observe that the fidelity of the surrogate improves as the number of counterfactuals increases. 
Besides, the performances of the surrogate model when the adversary does not use the architecture of the target model (Figures~\ref{subfig:unknown_public} and~\ref{subfig:unknown_private}) are slightly better than the performances of the surrogates trained using the same architecture as the target model (Figures~\ref{subfig:known_public} and~\ref{subfig:known_private}). 
These results also corroborate our previous findings that the target model architecture's knowledge does not provide a significant advantage to the adversary. 
Note that if the training data is used by the explanation API, this seems to give the adversary a small advantage in the lower query budget regime ($|Q_\adv{}| \leq 500$). 
However, in higher query budget regimes ($|Q_\adv{}| > 500$), it does not provide a significant advantage to the adversary.

\textbf{(S5) Impact of proximity and diversity on the performance of the model extraction attack.} Figure~\ref{fig:diverse_cf} summarizes the results obtained for scenario (S5) on the Adult Income dataset. 
Overall, the higher we set the constraints, the more likely the surrogate found will be of high fidelity. 
Similar to our previous observations, the knowledge of the target model architecture does not provide a significant advantage.

\textbf{Summary of the results.} 
Consistently across the experiments, we have observed that counterfactual explanations can be leveraged by an adversary with a limited query budget to perform high-fidelity and high-accuracy model extractions. 
In particular, when the adversary has partial or complete knowledge of the data distribution, he can obtain a high-fidelity and a high-accuracy surrogate with only $500$ queries. 
In contrast, when the data distribution is unknown, the surrogate performances are lower as expected. 
However, even in this restricted setting, the surrogate obtained with our attack still performs better than the surrogate generated using traditional model extraction attacks with full knowledge of the data distribution. 
Additionally, experiments with multiple and diverse counterfactuals demonstrate that this requirement leads to better performances of the model extraction attacks.

%% file: tables/bbox_summary.tex
\begin{table}[ht]
\centering
\begin{tabular}{@{}ccc@{}}
\toprule
\textbf{Dataset} & \textbf{Training Set} & \textbf{Test Set} \\ \midrule
Adult Income & $85.36$ & $84.70$ \\
COMPAS & $69.00$ & $66.30$ \\
Default Credit & $81.10$ & $80.70$ \\ \bottomrule
\end{tabular}%
\caption{\label{tab:bbox_summary} Performances of the black-box models. Columns report the accuracy of the black-box models on their training set and test set.}
\end{table}

%% file: tables/arch.tex
\begin{table*}[htbp]
\resizebox{\textwidth}{!}{%
\begin{tabular}{@{}cccccccc@{}}
\toprule
 & \textbf{Hidden layers} & \textbf{Hidden activation} & \textbf{Output activation} & \textbf{Loss} & \textbf{Optimizer} & \textbf{Regularizer} & \textbf{Epochs} \\ \midrule
\textbf{Arch 1} & $100,50$ & ReLu & Sigmoid & Binary cross-entropy & RMSprop & L$_1(0.001)$ & $100$ \\
\textbf{Arch 2} & $100,50$ & ReLu & Sigmoid & Binary cross-entropy & Adam & L$_1(0.01)$ & $20$ \\
\textbf{Arch 3} & $200,100,50,25$ & ReLu & Sigmoid & Binary cross-entropy & RMSprop & L$_1(0.01)$ & $20$ \\
\textbf{Arch 4} & $200,100,50,25$ & ReLu & Sigmoid & Binary cross-entropy & Adam & L$_1(0.01)$ & $20$ \\
\textbf{Arch 5} & $100,75,50$ & ReLu & Sigmoid & Binary cross-entropy & RMSprop & L$_1(0.001)$ & $100$ \\
\textbf{Arch 6} & $100,75,50$ & ReLu & Sigmoid & Binary cross-entropy & Adam & L$_1(0.01)$ & $20$ \\ \bottomrule
\end{tabular}%
}

\caption{\label{tab:arch} Architectures of the models used across the experiments. For both Adult Income and COMPAS datasets, we use \textbf{Arch 5} as the target model architecture, the adversary uses the remaining architectures as candidate architectures when the target model architecture is unknown. For the Default Credit dataset, \textbf{Arch 1} is used as the target model architecture and the remaining when the target model architecture is unknown.}
\end{table*}

%% file: tables/s1.tex
\begin{table*}[ht]
\resizebox{\textwidth}{!}{%
\begin{tabular}{@{}c|c|c|cccc|c@{}}
\toprule
\textbf{Dataset} & \textbf{\begin{tabular}[c]{@{}c@{}}Target model\\ Architecture\end{tabular}} & \textbf{\begin{tabular}[c]{@{}c@{}}Training data used \\ \\ by $\mathcal{E}(\cdot)$\end{tabular}} & \textbf{100 Queries} & \textbf{250 Queries} & \textbf{500 Queries} & \textbf{1000 Queries} & \textbf{\begin{tabular}[c]{@{}c@{}}Baseline\\ Model\end{tabular}} \\ \midrule
\multirow{4}{*}{Adult Income} & \multirow{2}{*}{known} & yes & $89.02/81.05$ & $92.06/82.94$ & $93.21/83.26$ & $94.22/83.68$ & \multirow{2}{*}{$81.28/76.06$} \\
 &  & no & $89.39/81.47$ & $91.78/82.87$ & $92.17/82.74$ & $94.84/83.88$ &  \\ \cmidrule(l){2-8} 
 & \multirow{2}{*}{unknown} & yes & $89.27/81.11$ & $92.42/83.18$ & $93.62/83.52$ & $94.65/83.88$ & \multirow{2}{*}{$81.28/76.06$} \\
 &  & no & $88.99/81.24$ & $92.21/83.05$ & $93.40/83.28$ & $94.89/83.97$ &  \\ \midrule
\multirow{4}{*}{COMPAS} & \multirow{2}{*}{known} & yes & $87.13/66.19$ & $91.29/67.30$ & $92.17/67.11$ & $92.85/66.97$ & \multirow{2}{*}{$71.42/61.09$} \\
 &  & no & $87.91/65.81$ & $89.57/65.86$ & $92.62/66.49$ & $93.92/66.50$ &  \\ \cmidrule(l){2-8} 
 & \multirow{2}{*}{unknown} & yes & $88.13/66.50$ & $90.81/67.26$ & $92.00/67.16$ & $92.36/66.90$ & \multirow{2}{*}{$85.04/64.95$} \\
 &  & no & $89.12/66.16$ & $90.08/66.03$ & $92.91/66.66$ & $93.43/66.49$ &  \\ \midrule
\multirow{4}{*}{Default Credit} & \multirow{2}{*}{known} & yes & $97.09/80.22$ & $97.93/80.55$ & $98.31/80.63$ & $98.57/80.52$ & \multirow{2}{*}{$88.52/77.86$} \\
 &  & no & $97.15/80.20$ & $97.77/80.34$ & $97.77/80.34$ & $98.28/80.48$ &  \\ \cmidrule(l){2-8} 
 & \multirow{2}{*}{unknown} & yes & $97.08/80.12$ & $97.99/80.57$ & $98.39/80.58$ & $98.39/80.58$ & \multirow{2}{*}{$88.52/77.86$} \\
 &  & no & $96.90/80.15$ & $97.52/80.38$ & $97.90/80.4$ & $98.03/80.43$ &  \\ \bottomrule
\end{tabular}%
}
\caption{\label{tab:s1} Performances (fidelity/accuracy) of the model extraction attack in scenario (S1) for Adult Income, COMPAS, and Default Credit datasets. For each of the query scenarios, we report on the performances (averaged over $10$ extraction attacks) of the surrogate model. The column of the baseline model correspond to the fidelity/accuracy of the surrogate model obtained using the whole attack pool $D_\adv{}$ to conduct a traditional model extraction attack.}
\end{table*}

%% file: tables/s2.tex
\begin{table*}[ht]
\resizebox{\textwidth}{!}{%
\begin{tabular}{@{}c|c|c|cccc@{}}
\toprule
\textbf{Dataset} & \textbf{\begin{tabular}[c]{@{}c@{}}Target model\\ Architecture\end{tabular}} & \textbf{\begin{tabular}[c]{@{}c@{}}Training data used \\ \\ by $\mathcal{E}(\cdot)$\end{tabular}} & \textbf{100 Queries} & \textbf{250 Queries} & \textbf{500 Queries} & \textbf{1000 Queries} \\ \midrule
\multirow{4}{*}{Adult Income} & \multirow{2}{*}{known} & yes & $86.19/79.47$ & $89.05/81.37$ & $91.70/82.84$ & $92.95/83.30$ \\
 &  & no & $86.48/79.82$ & $89.54/81.77$ & $91.74/82.84$ & $92.60/83.20$ \\ \cmidrule(l){2-7} 
 & \multirow{2}{*}{unknown} & yes & $86.22/79.46$ & $90.01/81.84$ & $92.14/83.12$ & $92.97/83.4$ \\
 &  & no & $86.22/79.83$ & $90.02/81.94$ & $92.13/83.09$ & $93.54/83.65$ \\ \bottomrule
\end{tabular}%
}

\caption{\label{tab:s2} Performances (fidelity/accuracy) of the model extraction attack in scenario (S2) for Adult Income. For each of the query scenarios, we report on the performances (averaged over $10$ extraction attacks) of the surrogate model.}
\end{table*}

%% file: tables/s3.tex
\begin{table*}[h!]
\resizebox{\textwidth}{!}{%
\begin{tabular}{@{}c|c|c|cccc@{}}
\toprule
\textbf{Dataset} & \textbf{\begin{tabular}[c]{@{}c@{}}Target model\\ Architecture\end{tabular}} & \textbf{\begin{tabular}[c]{@{}c@{}}Training data used \\ \\ by $\mathcal{E}(\cdot)$\end{tabular}} & \textbf{100 Queries} & \textbf{250 Queries} & \textbf{500 Queries} & \textbf{1000 Queries} \\ \midrule
\multirow{4}{*}{Adult Income} & \multirow{2}{*}{known} & yes & $82.30/75.90$ & $83.28/77.12$ & $84.46/78.25$ & $85.06/78.58$ \\
 &  & no & $82.31/76.11$ & $82.63/76.78$ & $83.74/77.57$ & $83.74/77.57$ \\ \cmidrule(l){2-7} 
 & \multirow{2}{*}{unknown} & yes & $81.98/75.48$ & $84.31/77.38$ & $85.75/78.79$ & $85.75/78.79$ \\
 &  & no & $81.58/75.59$ & $83.37/77.15$ & $84.60/78.28$ & $84.61/78.25$ \\ \bottomrule
\end{tabular}%
}

\caption{\label{tab:s3} Performances (fidelity/accuracy) of the model extraction attack in scenario (S3) for Adult Income. For each of the query scenarios, we report on the performances (averaged over $10$ extraction attacks) of the surrogate model.}
\end{table*}

%% file: related.tex
\section{Related work}
\label{sec:related}

As mentioned previously, model extraction attacks have been successfully conducted with the goal of obtaining high-accuracy and/or high-fidelity surrogates~\cite{tramer2016stealing,milli2019model,pal2019framework,correia2018copycat,papernot2017practical,orekondy2019knockoff,jagielski2020high} as well as with the objective to build functionally-equivalent surrogates~\cite{lowd2005adversarial,tramer2016stealing,batina2018csi,milli2019model,chandrasekaran2020exploring,jagielski2020high}. 

Most of the previous works aiming to build high-accuracy or high-fidelity surrogates usually rely on learning-based approaches. 
In this form of attack, the target model is used as an oracle to create a labelled dataset, which is then used as training data for the surrogates. 
Different learning-based approaches have been used, ranging from non-adaptive techniques (\emph{i.e.}, queries are sent independently from each other) as in~\cite{tramer2016stealing,correia2018copycat,milli2019model} to more advanced ones such as active learning~\cite{angluin1988queries} as in~\cite{tramer2016stealing,papernot2017practical,pal2019framework,chandrasekaran2020exploring} or semi-supervised learning~\cite{blum1998combining} as in~\cite{jagielski2020high}.
Our work falls into the first category as our attacks do not rely on adaptive techniques, which means, for instance, that all our queries could be sent at once in a batch. 
Despite being non-interactive, our attacks lead to high-fidelity and high-accuracy surrogates with low query budgets. 

Functionally-equivalent model extraction attacks often rely on equation-solving approaches~\cite{tramer2016stealing,jagielski2020high}, in which the adversary solves a system of equations modeling the unknown parameters of the target model to retrieve its weights. 
Advances techniques such as the use of power side-channel attacks~\cite{batina2018csi} and gradients-based explanations~\cite{milli2019model} have been used to improve the performances of functionally-equivalent model extraction attacks. 

In addition, hybrid approaches~\cite{jagielski2020high} combining functionally-equivalent attacks and learning-based attacks have been used to improve the overall performance of the surrogate model. 
Usually, these approaches first fix some unknown model parameters to the values obtained through a functionally-equivalent attack to reduce the number of free variables before training a surrogate through a learning-based attack by leveraging the parameters extracted during the first phase. 
Other works have investigated how to improve the deployment knowledge of the adversary such as~\cite{wang2018stealing}, in which the authors propose an attack to steal the hyperparameters of black-box models and~\cite{oh2019towards}, in which the authors have designed attacks to infer the architecture as well as training hyperparameters of black-box ML models.  

With respect to designing model extraction attacks that leverage explanations provided by the model, to the best of our knowledge, there exists currently only one work~\cite{milli2019model} (which does not rely at all on counterfactual explanations). 
In this seminal paper, the authors demonstrated that an adversary could perform a model extraction attack by relying on the target model's gradient explanations. 
Gradient explanations are used in visualization-based explanation techniques~\cite{baehrens2010explain,simonyan2013deep} to highlight parts of an image that lead to the decision of the target model. 
However, they are challenging to interpret in prediction problems involving tabular data, which are the setting in which we focus on in this paper. 
In addition, our technique is agnostic to the architecture of the target model, while the attack in~\cite{milli2019model} is designed for a special family of models, namely $2$-layer neural networks.

%% file: discussions.tex
\section{Discussion}
\label{sec:discussions}

In this section, we discuss the possible countermeasures that could be deployed to mitigate our attacks as well as the inherent existing tension between the two requirements that are privacy and explainability.

\textbf{Countermeasures.} 
Protection mechanisms against model extraction can be categorized into two categories: defenses that aim to prevent theft-motivated model extraction attacks and those that can be used to prevent the adversary from learning a high-fidelity surrogate. 
In the former case, since the adversary is motivated by stealing the model for its own benefit (\emph{e.g.}, by deploying it as a MLAAS), he can be deterred to do so through defenses techniques mainly based on embedding watermarks in the surrogate model~\cite{szyller2019dawn,jia2020entangled,le2020adversarial}. Such watermarks can then be detected if the adversary later makes the surrogate model publicly available for queries. 
However, watermark-based techniques are inefficient against adversaries that use stolen models internally.

For the prevention of high-fidelity model extraction attacks, the defense mechanisms proposed usually rely on query monitoring and auditing techniques that analyze the query pattern to distinguish normal users from adversaries~\cite{kesarwani2018model,juuti2019prada,atli2019extraction}. 
However, such approaches will be inefficient against non-adaptive attacks that work with low query budgets, such as the attack we proposed, because it will be very difficult to distinguish adversarial queries from the ones made by regular users. 
In addition, an adversary can always perform a Sybil attack in which he creates multiple accounts under different identities, before sharing his queries among these ``regular''-looking users to avoid detection.

\textbf{Tension between privacy and explainability.} 
Post-hoc explanation techniques are often presented as a way to fulfill two distinct objectives. 
On the one hand, they can be used as debugging tools to inform experts such as data scientists or machine learning researchers on the behavior of their black-box models. 
On the other hand, they can be used as justification to explain the outcomes of deployed black-box models to the end users~\cite{hancox2020robustness}. 
The requirements that are asked from post-hoc explanations are very different depending on which of these two settings is considered. 
For instance, to be acceptable in the second context, post-hoc explanations need to be realistic by satisfying criteria such as robustness and diversity. 
At the same time, the more realistic they become, the more information they will leak about the black-box model they are explaining, which will lead to more powerful attacks. 
In particular, our paper demonstrates how an adversary can leverage counterfactual explanations to devise high-fidelity and high-accuracy model extraction attacks. 
In addition, the performances of the surrogate models can only increase as post-hoc explanations get more realistic.
Thus, we believe that often there will be a trade-off to set between the realism of explanations and the privacy protection that we aim at achieving against model extraction or other privacy attacks on machine learning models.

A recent work has suggested the use of differential privacy~\cite{dwork2008differential,dwork2006calibrating} in the design of post-hoc explanations~\cite{patel2020model}. 
However, the impact of differential privacy on the robustness of the explanations and the trust we can have on differentially-private explanations remains an open question. 
For high-stake decision systems, it seems that the safer solution would be to directly design an inherently transparent models built in a privacy-preserving way by using techniques such as model-agnostic private learning~\cite{papernotAEGT17,bassily2018model}.

%% file: conclusion.tex
\section{Conclusion}
\label{sec:conclusion}

In this work, we have investigated counterfactual explanations-based model extraction attacks in five different adversarial scenarios. 
In particular, we have demonstrated that an adversary can exploit counterfactual explanations to conduct high-fidelity and high-accuracy learning-based model extraction attacks even under low query budgets. 
Furthermore, if the counterfactual explanations provided are required to be diverse to increase the trust in the explanations provided, then this directly improves the performances of the surrogate models learnt by our attacks.

As shown by previous works on fairwashing~\cite{aivodji2019fairwashing,fukuchi2019pretending,slack2019can}, post-hoc explanations techniques are vulnerable to explanation manipulations since they can be unfaithful to the black-box model they are explaining (\emph{e.g.}, by giving the impression that the model is fair while it is not the case). 
On the other hand, our work demonstrates that the more faithful ML models get, the more powerful are the model extraction attacks that an adversary can perform. 
Future work will investigate the use of privacy-preserving transparent box design as a solution to solve the tension between privacy and explainability.

%% file: main.bbl
\begin{thebibliography}{10}
\providecommand{\url}[1]{#1}
\csname url@samestyle\endcsname
\providecommand{\newblock}{\relax}
\providecommand{\bibinfo}[2]{#2}
\providecommand{\BIBentrySTDinterwordspacing}{\spaceskip=0pt\relax}
\providecommand{\BIBentryALTinterwordstretchfactor}{4}
\providecommand{\BIBentryALTinterwordspacing}{\spaceskip=\fontdimen2\font plus
\BIBentryALTinterwordstretchfactor\fontdimen3\font minus
  \fontdimen4\font\relax}
\providecommand{\BIBforeignlanguage}[2]{{%
\expandafter\ifx\csname l@#1\endcsname\relax
\typeout{** WARNING: IEEEtran.bst: No hyphenation pattern has been}%
\typeout{** loaded for the language `#1'. Using the pattern for}%
\typeout{** the default language instead.}%
\else
\language=\csname l@#1\endcsname
\fi
#2}}
\providecommand{\BIBdecl}{\relax}
\BIBdecl

\bibitem{siddiqi2012credit}
N.~Siddiqi, \emph{Credit risk scorecards: developing and implementing
  intelligent credit scoring}.\hskip 1em plus 0.5em minus 0.4em\relax John
  Wiley \& Sons, 2012, vol.~3.

\bibitem{kleinberg2017human}
J.~Kleinberg, H.~Lakkaraju, J.~Leskovec, J.~Ludwig, and S.~Mullainathan,
  ``Human decisions and machine predictions,'' \emph{The quarterly journal of
  economics}, vol. 133, no.~1, pp. 237--293, 2017.

\bibitem{miller_2015}
C.~C. Miller, ``Can an algorithm hire better than a human?'' Jun 2015.

\bibitem{wexler2017computer}
R.~Wexler, ``When a computer program keeps you in jail: How computers are
  harming criminal justice,'' \emph{New York Times}, 2017.

\bibitem{floridi2019unified}
L.~Floridi and J.~Cowls, ``A unified framework of five principles for ai in
  society,'' \emph{Harvard Data Science Review}, 2019.

\bibitem{jobin2019global}
A.~Jobin, M.~Ienca, and E.~Vayena, ``The global landscape of ai ethics
  guidelines,'' \emph{Nature Machine Intelligence}, pp. 1--11, 2019.

\bibitem{goodman2016european}
B.~Goodman and S.~Flaxman, ``European union regulations on algorithmic
  decision-making and a ``right to explanation'','' \emph{AI Magazine},
  vol.~38, no.~3, pp. 50--57, 2017.

\bibitem{lipton2016mythos}
Z.~C. Lipton, ``The mythos of model interpretability,'' \emph{Communications of
  the {ACM}}, vol.~61, no.~10, pp. 36--43, 2018.

\bibitem{lepri2017fair}
B.~Lepri, N.~Oliver, E.~Letouz{\'e}, A.~Pentland, and P.~Vinck, ``Fair,
  transparent, and accountable algorithmic decision-making processes,''
  \emph{Philosophy \& Technology}, pp. 1--17, 2017.

\bibitem{montavon2018methods}
G.~Montavon, W.~Samek, and K.-R. M{\"u}ller, ``Methods for interpreting and
  understanding deep neural networks,'' \emph{Digital Signal Processing},
  vol.~73, pp. 1--15, 2018.

\bibitem{guidotti2019survey}
R.~Guidotti, A.~Monreale, S.~Ruggieri, F.~Turini, F.~Giannotti, and
  D.~Pedreschi, ``A survey of methods for explaining black box models,''
  \emph{ACM computing surveys (CSUR)}, vol.~51, no.~5, p.~93, 2019.

\bibitem{arrieta2019explainable}
A.~B. Arrieta, N.~D{\'\i}az-Rodr{\'\i}guez, J.~Del~Ser, A.~Bennetot, S.~Tabik,
  A.~Barbado, S.~Garc{\'\i}a, S.~Gil-L{\'o}pez, D.~Molina, R.~Benjamins
  \emph{et~al.}, ``Explainable artificial intelligence (xai): Concepts,
  taxonomies, opportunities and challenges toward responsible ai,'' \emph{arXiv
  preprint arXiv:1910.10045}, 2019.

\bibitem{li2002mining}
J.~Li, H.~Shen, and R.~Topor, ``Mining the optimal class association rule
  set,'' \emph{Knowledge-Based Systems}, vol.~15, no.~7, pp. 399--405, 2002.

\bibitem{angelino2017learning}
E.~Angelino, N.~Larus-Stone, D.~Alabi, M.~Seltzer, and C.~Rudin, ``Learning
  certifiably optimal rule lists,'' in \emph{Proceedings of the 23rd ACM SIGKDD
  International Conference on Knowledge Discovery and Data Mining}.\hskip 1em
  plus 0.5em minus 0.4em\relax Halifax, NS, Canada: ACM, 2017, pp. 35--44.

\bibitem{breiman2017classification}
L.~Breiman, \emph{Classification and regression trees}.\hskip 1em plus 0.5em
  minus 0.4em\relax Routledge, 2017.

\bibitem{ustun2016supersparse}
B.~Ustun and C.~Rudin, ``Supersparse linear integer models for optimized
  medical scoring systems,'' \emph{Machine Learning}, vol. 102, no.~3, pp.
  349--391, 2016.

\bibitem{rijnbeek2010finding}
P.~R. Rijnbeek and J.~A. Kors, ``Finding a short and accurate decision rule in
  disjunctive normal form by exhaustive search,'' \emph{Machine learning},
  vol.~80, no.~1, pp. 33--62, 2010.

\bibitem{mccormick2011hierarchical}
T.~McCormick, C.~Rudin, and D.~Madigan, ``A hierarchical model for association
  rule mining of sequential events: An approach to automated medical symptom
  prediction,'' 2011.

\bibitem{dash2018boolean}
S.~Dash, O.~Gunluk, and D.~Wei, ``Boolean decision rules via column
  generation,'' in \emph{Advances in Neural Information Processing Systems},
  2018, pp. 4655--4665.

\bibitem{yang2017scalable}
H.~Yang, C.~Rudin, and M.~Seltzer, ``Scalable bayesian rule lists,'' in
  \emph{Proceedings of the 34th International Conference on Machine
  Learning-Volume 70}.\hskip 1em plus 0.5em minus 0.4em\relax JMLR. org, 2017,
  pp. 3921--3930.

\bibitem{wang2015falling}
F.~Wang and C.~Rudin, ``Falling rule lists,'' in \emph{Artificial Intelligence
  and Statistics}, 2015, pp. 1013--1022.

\bibitem{aivodji2019learning}
U.~A{\"\i}vodji, J.~Ferry, S.~Gambs, M.-J. Huguet, and M.~Siala, ``Learning
  fair rule lists,'' \emph{arXiv preprint arXiv:1909.03977}, 2019.

\bibitem{narodytska2018learning}
N.~Narodytska, A.~Ignatiev, F.~Pereira, J.~Marques-Silva, and I.~RAS,
  ``Learning optimal decision trees with sat.'' in \emph{IJCAI}, 2018, pp.
  1362--1368.

\bibitem{zeng2017interpretable}
J.~Zeng, B.~Ustun, and C.~Rudin, ``Interpretable classification models for
  recidivism prediction,'' \emph{Journal of the Royal Statistical Society:
  Series A (Statistics in Society)}, vol. 180, no.~3, pp. 689--722, 2017.

\bibitem{koh2006two}
H.~C. Koh, W.~C. Tan, and C.~P. Goh, ``A two-step method to construct credit
  scoring models with data mining techniques,'' \emph{International Journal of
  Business and Information}, vol.~1, no.~1, 2006.

\bibitem{craven1996extracting}
M.~Craven and J.~W. Shavlik, ``Extracting tree-structured representations of
  trained networks,'' in \emph{Advances in neural information processing
  systems}, 1996, pp. 24--30.

\bibitem{ribeiro2016should}
M.~T. Ribeiro, S.~Singh, and C.~Guestrin, ``Why should {I} trust you?:
  Explaining the predictions of any classifier,'' in \emph{Proceedings of the
  22nd ACM SIGKDD International Conference on Knowledge Discovery and Data
  Mining (KDD'16)}.\hskip 1em plus 0.5em minus 0.4em\relax ACM, 2016, pp.
  1135--1144.

\bibitem{lundberg2017unified}
S.~M. Lundberg and S.-I. Lee, ``A unified approach to interpreting model
  predictions,'' in \emph{Proceedings of the Annual Conference on Neural
  Information Processing Systems (NIPS'17)}, 2017, pp. 4765--4774.

\bibitem{cortez2013using}
P.~Cortez and M.~J. Embrechts, ``Using sensitivity analysis and visualization
  techniques to open black box data mining models,'' \emph{Information
  Sciences}, vol. 225, pp. 1--17, 2013.

\bibitem{krause2016interacting}
J.~Krause, A.~Perer, and K.~Ng, ``Interacting with predictions: Visual
  inspection of black-box machine learning models,'' in \emph{Proceedings of
  the 2016 CHI Conference on Human Factors in Computing Systems}, 2016, pp.
  5686--5697.

\bibitem{erhan2009visualizing}
D.~Erhan, Y.~Bengio, A.~Courville, and P.~Vincent, ``Visualizing higher-layer
  features of a deep network,'' \emph{University of Montreal}, vol. 1341,
  no.~3, p.~1, 2009.

\bibitem{baehrens2010explain}
D.~Baehrens, T.~Schroeter, S.~Harmeling, M.~Kawanabe, K.~Hansen, and K.-R.
  M{\"u}ller, ``How to explain individual classification decisions,'' \emph{The
  Journal of Machine Learning Research}, vol.~11, pp. 1803--1831, 2010.

\bibitem{simonyan2013deep}
K.~Simonyan, A.~Vedaldi, and A.~Zisserman, ``Deep inside convolutional
  networks: Visualising image classification models and saliency maps,''
  \emph{arXiv preprint arXiv:1312.6034}, 2013.

\bibitem{kim2016examples}
B.~Kim, R.~Khanna, and O.~O. Koyejo, ``Examples are not enough, learn to
  criticize! criticism for interpretability,'' in \emph{Advances in neural
  information processing systems}, 2016, pp. 2280--2288.

\bibitem{wachter2017counterfactual}
S.~Wachter, B.~Mittelstadt, and C.~Russell, ``Counterfactual explanations
  without opening the black box: Automated decisions and the gdpr,''
  \emph{Harv. JL \& Tech.}, vol.~31, p. 841, 2017.

\bibitem{adebayo2018sanity}
J.~Adebayo, J.~Gilmer, M.~Muelly, I.~Goodfellow, M.~Hardt, and B.~Kim, ``Sanity
  checks for saliency maps,'' in \emph{Advances in Neural Information
  Processing Systems}, 2018, pp. 9505--9515.

\bibitem{rudin2019stop}
C.~Rudin, ``Stop explaining black box machine learning models for high stakes
  decisions and use interpretable models instead,'' \emph{Nature Machine
  Intelligence}, vol.~1, no.~5, pp. 206--215, 2019.

\bibitem{aivodji2019fairwashing}
U.~A{\"\i}vodji, H.~Arai, O.~Fortineau, S.~Gambs, S.~Hara, and A.~Tapp,
  ``Fairwashing: the risk of rationalization,'' in \emph{International
  Conference on Machine Learning}, 2019, pp. 161--170.

\bibitem{fukuchi2019pretending}
K.~Fukuchi, S.~Hara, and T.~Maehara, ``Pretending fair decisions via stealthily
  biased sampling,'' \emph{arXiv preprint arXiv:1901.08291}, 2019.

\bibitem{laugel2019dangers}
T.~Laugel, M.-J. Lesot, C.~Marsala, X.~Renard, and M.~Detyniecki, ``The dangers
  of post-hoc interpretability: unjustified counterfactual explanations,'' in
  \emph{Proceedings of the 28th International Joint Conference on Artificial
  Intelligence}.\hskip 1em plus 0.5em minus 0.4em\relax AAAI Press, 2019, pp.
  2801--2807.

\bibitem{ghorbani2019interpretation}
A.~Ghorbani, A.~Abid, and J.~Zou, ``Interpretation of neural networks is
  fragile,'' in \emph{Proceedings of the AAAI Conference on Artificial
  Intelligence}, vol.~33, 2019, pp. 3681--3688.

\bibitem{heo2019fooling}
J.~Heo, S.~Joo, and T.~Moon, ``Fooling neural network interpretations via
  adversarial model manipulation,'' \emph{arXiv preprint arXiv:1902.02041},
  2019.

\bibitem{dombrowski2019explanations}
A.-K. Dombrowski, M.~Alber, C.~J. Anders, M.~Ackermann, K.-R. M{\"u}ller, and
  P.~Kessel, ``Explanations can be manipulated and geometry is to blame,''
  \emph{arXiv preprint arXiv:1906.07983}, 2019.

\bibitem{merrer2019bouncer}
E.~L. Merrer and G.~Tredan, ``The bouncer problem: Challenges to remote
  explainability,'' \emph{arXiv preprint arXiv:1910.01432}, 2019.

\bibitem{lakkaraju2019fool}
H.~Lakkaraju and O.~Bastani, ``" how do i fool you?": Manipulating user trust
  via misleading black box explanations,'' \emph{arXiv preprint
  arXiv:1911.06473}, 2019.

\bibitem{slack2019can}
D.~Slack, S.~Hilgard, E.~Jia, S.~Singh, and H.~Lakkaraju, ``How can we fool
  lime and shap? adversarial attacks on post hoc explanation methods,''
  \emph{arXiv preprint arXiv:1911.02508}, 2019.

\bibitem{zhang2020interpretable}
X.~Zhang, N.~Wang, H.~Shen, S.~Ji, X.~Luo, and T.~Wang, ``Interpretable deep
  learning under fire,'' in \emph{29th $\{$USENIX$\}$ Security Symposium
  ($\{$USENIX$\}$ Security 20)}, 2020.

\bibitem{shokri2019privacy}
R.~Shokri, M.~Strobel, and Y.~Zick, ``Privacy risks of explaining machine
  learning models,'' \emph{arXiv preprint arXiv:1907.00164}, 2019.

\bibitem{milli2019model}
S.~Milli, L.~Schmidt, A.~D. Dragan, and M.~Hardt, ``Model reconstruction from
  model explanations,'' in \emph{Proceedings of the Conference on Fairness,
  Accountability, and Transparency}, 2019, pp. 1--9.

\bibitem{murphy2012machine}
K.~P. Murphy, \emph{Machine learning: a probabilistic perspective}.\hskip 1em
  plus 0.5em minus 0.4em\relax MIT press, 2012.

\bibitem{breiman2001random}
L.~Breiman, ``Random forests,'' \emph{Machine learning}, vol.~45, no.~1, pp.
  5--32, 2001.

\bibitem{rivest1987learning}
R.~L. Rivest, ``Learning decision lists,'' \emph{Machine learning}, vol.~2,
  no.~3, pp. 229--246, 1987.

\bibitem{Goodfellow-et-al-2016}
I.~Goodfellow, Y.~Bengio, and A.~Courville, \emph{Deep Learning}.\hskip 1em
  plus 0.5em minus 0.4em\relax MIT Press, 2016.

\bibitem{frank2010uci}
A.~Frank and A.~Asuncion, ``Uci machine learning repository [http://archive.
  ics. uci. edu/ml]. irvine, ca: University of california,'' \emph{School of
  information and computer science}, vol. 213, pp. 2--2, 2010.

\bibitem{lash2017generalized}
M.~T. Lash, Q.~Lin, N.~Street, J.~G. Robinson, and J.~Ohlmann, ``Generalized
  inverse classification,'' in \emph{Proceedings of the 2017 SIAM International
  Conference on Data Mining}.\hskip 1em plus 0.5em minus 0.4em\relax SIAM,
  2017, pp. 162--170.

\bibitem{laugel2017inverse}
T.~Laugel, M.-J. Lesot, C.~Marsala, X.~Renard, and M.~Detyniecki, ``Inverse
  classification for comparison-based interpretability in machine learning,''
  \emph{arXiv preprint arXiv:1712.08443}, 2017.

\bibitem{tolomei2017interpretable}
G.~Tolomei, F.~Silvestri, A.~Haines, and M.~Lalmas, ``Interpretable predictions
  of tree-based ensembles via actionable feature tweaking,'' in
  \emph{Proceedings of the 23rd ACM SIGKDD international conference on
  knowledge discovery and data mining}, 2017, pp. 465--474.

\bibitem{grath2018interpretable}
R.~M. Grath, L.~Costabello, C.~L. Van, P.~Sweeney, F.~Kamiab, Z.~Shen, and
  F.~Lecue, ``Interpretable credit application predictions with counterfactual
  explanations,'' \emph{arXiv preprint arXiv:1811.05245}, 2018.

\bibitem{russell2019efficient}
C.~Russell, ``Efficient search for diverse coherent explanations,'' in
  \emph{Proceedings of the Conference on Fairness, Accountability, and
  Transparency}, 2019, pp. 20--28.

\bibitem{ustun2019actionable}
B.~Ustun, A.~Spangher, and Y.~Liu, ``Actionable recourse in linear
  classification,'' in \emph{Proceedings of the Conference on Fairness,
  Accountability, and Transparency}, 2019, pp. 10--19.

\bibitem{joshi2019towards}
S.~Joshi, O.~Koyejo, W.~Vijitbenjaronk, B.~Kim, and J.~Ghosh, ``Towards
  realistic individual recourse and actionable explanations in black-box
  decision making systems,'' \emph{arXiv preprint arXiv:1907.09615}, 2019.

\bibitem{pawelczyk2020learning}
M.~Pawelczyk, K.~Broelemann, and G.~Kasneci, ``Learning model-agnostic
  counterfactual explanations for tabular data,'' in \emph{Proceedings of The
  Web Conference 2020}, 2020, pp. 3126--3132.

\bibitem{mothilal2020explaining}
R.~K. Mothilal, A.~Sharma, and C.~Tan, ``Explaining machine learning
  classifiers through diverse counterfactual explanations,'' in
  \emph{Proceedings of the 2020 Conference on Fairness, Accountability, and
  Transparency}, 2020, pp. 607--617.

\bibitem{karimi2020model}
A.-H. Karimi, G.~Barthe, B.~Balle, and I.~Valera, ``Model-agnostic
  counterfactual explanations for consequential decisions,'' in
  \emph{International Conference on Artificial Intelligence and Statistics},
  2020, pp. 895--905.

\bibitem{atli2019extraction}
B.~G. Atli, S.~Szyller, M.~Juuti, S.~Marchal, and N.~Asokan, ``Extraction of
  complex dnn models: Real threat or boogeyman?'' \emph{arXiv preprint
  arXiv:1910.05429}, 2019.

\bibitem{jagielski2020high}
M.~Jagielski, N.~Carlini, D.~Berthelot, A.~Kurakin, and N.~Papernot, ``High
  accuracy and high fidelity extraction of neural networks,'' in \emph{29th
  $\{$USENIX$\}$ Security Symposium ($\{$USENIX$\}$ Security 20)}, 2020.

\bibitem{fredrikson2014privacy}
M.~Fredrikson, E.~Lantz, S.~Jha, S.~Lin, D.~Page, and T.~Ristenpart, ``Privacy
  in pharmacogenetics: An end-to-end case study of personalized warfarin
  dosing,'' in \emph{23rd $\{$USENIX$\}$ Security Symposium ($\{$USENIX$\}$
  Security 14)}, 2014, pp. 17--32.

\bibitem{fredrikson2015model}
M.~Fredrikson, S.~Jha, and T.~Ristenpart, ``Model inversion attacks that
  exploit confidence information and basic countermeasures,'' in
  \emph{Proceedings of the 22nd ACM SIGSAC Conference on Computer and
  Communications Security}, 2015, pp. 1322--1333.

\bibitem{szegedy2013intriguing}
C.~Szegedy, W.~Zaremba, I.~Sutskever, J.~Bruna, D.~Erhan, I.~Goodfellow, and
  R.~Fergus, ``Intriguing properties of neural networks,'' \emph{arXiv preprint
  arXiv:1312.6199}, 2013.

\bibitem{goodfellow2014explaining}
I.~J. Goodfellow, J.~Shlens, and C.~Szegedy, ``Explaining and harnessing
  adversarial examples,'' \emph{arXiv preprint arXiv:1412.6572}, 2014.

\bibitem{papernot2017practical}
N.~Papernot, P.~McDaniel, I.~Goodfellow, S.~Jha, Z.~B. Celik, and A.~Swami,
  ``Practical black-box attacks against machine learning,'' in
  \emph{Proceedings of the 2017 ACM on Asia conference on computer and
  communications security}, 2017, pp. 506--519.

\bibitem{angwin2016machine}
J.~Angwin, J.~Larson, S.~Mattu, and L.~Kirchner, ``Machine bias,''
  \emph{ProPublica, May}, vol.~23, 2016.

\bibitem{tieleman2012lecture}
T.~Tieleman and G.~Hinton, ``Lecture 6.5-rmsprop: Divide the gradient by a
  running average of its recent magnitude,'' \emph{COURSERA: Neural networks
  for machine learning}, vol.~4, no.~2, pp. 26--31, 2012.

\bibitem{elsken2019neural}
T.~Elsken, J.~H. Metzen, and F.~Hutter, ``Neural architecture search: A
  survey,'' \emph{Journal of Machine Learning Research}, vol.~20, pp. 1--21,
  2019.

\bibitem{tramer2016stealing}
F.~Tram{\`e}r, F.~Zhang, A.~Juels, M.~K. Reiter, and T.~Ristenpart, ``Stealing
  machine learning models via prediction apis,'' in \emph{25th $\{$USENIX$\}$
  Security Symposium ($\{$USENIX$\}$ Security 16)}, 2016, pp. 601--618.

\bibitem{pal2019framework}
S.~Pal, Y.~Gupta, A.~Shukla, A.~Kanade, S.~Shevade, and V.~Ganapathy, ``A
  framework for the extraction of deep neural networks by leveraging public
  data,'' \emph{arXiv preprint arXiv:1905.09165}, 2019.

\bibitem{correia2018copycat}
J.~R. Correia-Silva, R.~F. Berriel, C.~Badue, A.~F. de~Souza, and
  T.~Oliveira-Santos, ``Copycat cnn: Stealing knowledge by persuading
  confession with random non-labeled data,'' in \emph{2018 International Joint
  Conference on Neural Networks (IJCNN)}.\hskip 1em plus 0.5em minus
  0.4em\relax IEEE, 2018, pp. 1--8.

\bibitem{orekondy2019knockoff}
T.~Orekondy, B.~Schiele, and M.~Fritz, ``Knockoff nets: Stealing functionality
  of black-box models,'' in \emph{Proceedings of the IEEE Conference on
  Computer Vision and Pattern Recognition}, 2019, pp. 4954--4963.

\bibitem{lowd2005adversarial}
D.~Lowd and C.~Meek, ``Adversarial learning,'' in \emph{Proceedings of the
  eleventh ACM SIGKDD international conference on Knowledge discovery in data
  mining}, 2005, pp. 641--647.

\bibitem{batina2018csi}
L.~Batina, S.~Bhasin, D.~Jap, and S.~Picek, ``Csi neural network: Using
  side-channels to recover your artificial neural network information,''
  \emph{arXiv preprint arXiv:1810.09076}, 2018.

\bibitem{chandrasekaran2020exploring}
V.~Chandrasekaran, K.~Chaudhuri, I.~Giacomelli, S.~Jha, and S.~Yan, ``Exploring
  connections between active learning and model extraction,'' in \emph{29th
  $\{$USENIX$\}$ Security Symposium ($\{$USENIX$\}$ Security 20)}, 2020, pp.
  1309--1326.

\bibitem{angluin1988queries}
D.~Angluin, ``Queries and concept learning,'' \emph{Machine learning}, vol.~2,
  no.~4, pp. 319--342, 1988.

\bibitem{blum1998combining}
A.~Blum and T.~Mitchell, ``Combining labeled and unlabeled data with
  co-training,'' in \emph{Proceedings of the eleventh annual conference on
  Computational learning theory}, 1998, pp. 92--100.

\bibitem{wang2018stealing}
B.~Wang and N.~Z. Gong, ``Stealing hyperparameters in machine learning,'' in
  \emph{2018 IEEE Symposium on Security and Privacy (SP)}.\hskip 1em plus 0.5em
  minus 0.4em\relax IEEE, 2018, pp. 36--52.

\bibitem{oh2019towards}
S.~J. Oh, B.~Schiele, and M.~Fritz, ``Towards reverse-engineering black-box
  neural networks,'' in \emph{Explainable AI: Interpreting, Explaining and
  Visualizing Deep Learning}.\hskip 1em plus 0.5em minus 0.4em\relax Springer,
  2019, pp. 121--144.

\bibitem{szyller2019dawn}
S.~Szyller, B.~G. Atli, S.~Marchal, and N.~Asokan, ``Dawn: Dynamic adversarial
  watermarking of neural networks,'' \emph{arXiv preprint arXiv:1906.00830},
  2019.

\bibitem{jia2020entangled}
H.~Jia, C.~A. Choquette-Choo, and N.~Papernot, ``Entangled watermarks as a
  defense against model extraction,'' \emph{arXiv preprint arXiv:2002.12200},
  2020.

\bibitem{le2020adversarial}
E.~Le~Merrer, P.~Perez, and G.~Tr{\'e}dan, ``Adversarial frontier stitching for
  remote neural network watermarking,'' \emph{Neural Computing and
  Applications}, vol.~32, no.~13, pp. 9233--9244, 2020.

\bibitem{kesarwani2018model}
M.~Kesarwani, B.~Mukhoty, V.~Arya, and S.~Mehta, ``Model extraction warning in
  mlaas paradigm,'' in \emph{Proceedings of the 34th Annual Computer Security
  Applications Conference}, 2018, pp. 371--380.

\bibitem{juuti2019prada}
M.~Juuti, S.~Szyller, S.~Marchal, and N.~Asokan, ``Prada: protecting against
  dnn model stealing attacks,'' in \emph{2019 IEEE European Symposium on
  Security and Privacy (EuroS\&P)}.\hskip 1em plus 0.5em minus 0.4em\relax
  IEEE, 2019, pp. 512--527.

\bibitem{hancox2020robustness}
L.~Hancox-Li, ``Robustness in machine learning explanations: does it matter?''
  in \emph{Proceedings of the 2020 Conference on Fairness, Accountability, and
  Transparency}, 2020, pp. 640--647.

\bibitem{dwork2008differential}
C.~Dwork, ``Differential privacy: A survey of results,'' in \emph{International
  conference on theory and applications of models of computation}.\hskip 1em
  plus 0.5em minus 0.4em\relax Springer, 2008, pp. 1--19.

\bibitem{dwork2006calibrating}
C.~Dwork, F.~McSherry, K.~Nissim, and A.~Smith, ``Calibrating noise to
  sensitivity in private data analysis,'' in \emph{Theory of cryptography
  conference}.\hskip 1em plus 0.5em minus 0.4em\relax Springer, 2006, pp.
  265--284.

\bibitem{patel2020model}
N.~Patel, R.~Shokri, and Y.~Zick, ``Model explanations with differential
  privacy,'' \emph{arXiv preprint arXiv:2006.09129}, 2020.

\bibitem{papernotAEGT17}
N.~Papernot, M.~Abadi, {\'{U}}.~Erlingsson, I.~J. Goodfellow, and K.~Talwar,
  ``Semi-supervised knowledge transfer for deep learning from private training
  data,'' in \emph{5th International Conference on Learning Representations,
  {ICLR} 2017, Toulon, France, April 24-26, 2017, Conference Track
  Proceedings}.\hskip 1em plus 0.5em minus 0.4em\relax OpenReview.net, 2017.

\bibitem{bassily2018model}
R.~Bassily, O.~Thakkar, and A.~G. Thakurta, ``Model-agnostic private
  learning,'' in \emph{Advances in Neural Information Processing Systems},
  2018, pp. 7102--7112.

\end{thebibliography}
